\documentclass[final]{cvpr}
\usepackage{times}
\usepackage{epsfig}
\usepackage{graphicx}
\usepackage{amsmath}
\usepackage{amssymb}
\usepackage{mathrsfs}
\usepackage{mathtools, cuted}
\usepackage{lipsum, color}
\usepackage{bbding}
\usepackage{cuted}
\usepackage{capt-of}
\usepackage{multirow}
\usepackage{makecell}
\usepackage{appendix}
\usepackage{multicol}
\usepackage{overpic}
\usepackage[ruled,lined,boxed,commentsnumbered]{algorithm2e}

\usepackage[pagebackref=true,breaklinks=true,colorlinks,bookmarks=false]{hyperref}
\def\cvprPaperID{7219} 
\def\confYear{CVPR 2021}

\begin{document}
\title{ConvTransformer: A Convolutional Transformer Network for Video Frame Synthesis}
\author{Zhouyong Liu \and Shun Luo \and Wubin Li \and Jingben Lu \and Yufan Wu \and Shilei Sun \and Chunguo Li \and {Luxi Yang \thanks{Corresponding author}}  \and
School of Information Science and Engineering, Southeast University}

\makeatletter
\let\@oldmaketitle\@maketitle 
\renewcommand{\@maketitle}{\@oldmaketitle
  \centering
  \includegraphics[width=0.95\linewidth]{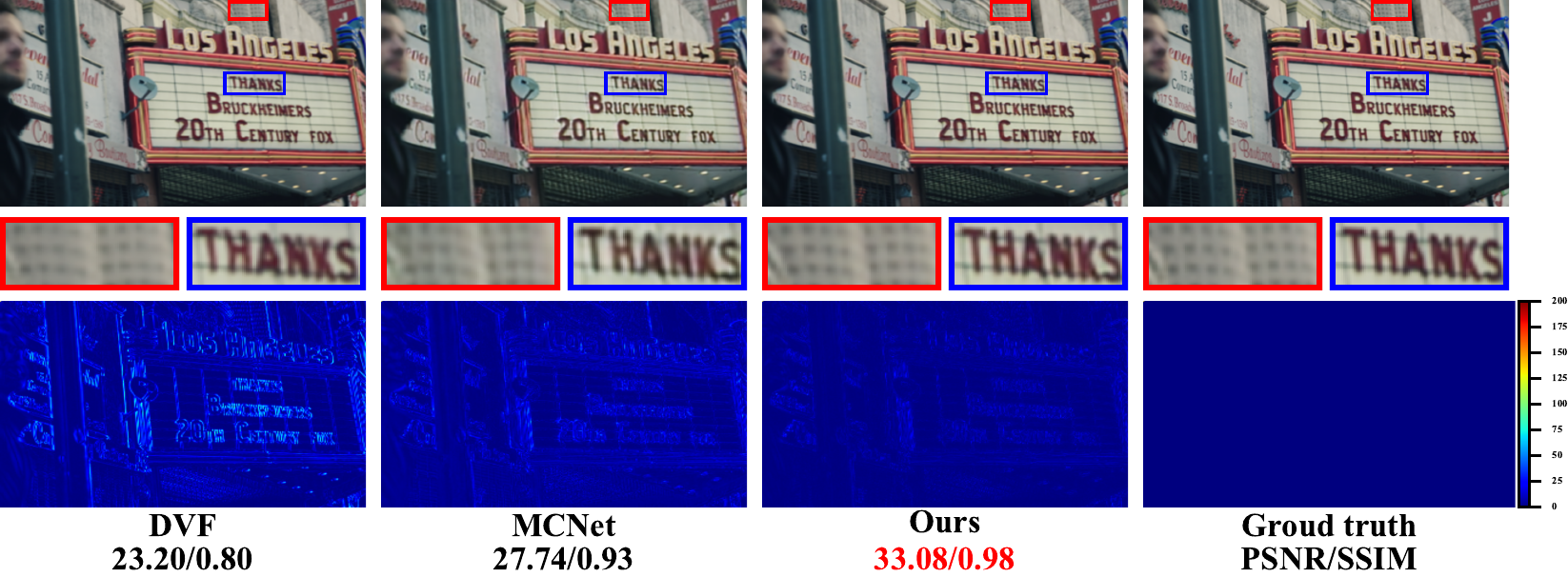}
  \captionof{figure}{\textbf{Example of video frame extrapolation.} Top is the extrapolated result, middle is the zoomed local details and bottom is the occlusion map that depicts the residuals between extrapolated image and ground truth. ColorBar with gradient from \textcolor{blue}{blue} to \textcolor{red}{red} indicates the residual intensity in the range 0 to 255. The zoomed-in details indicate that ConvTransformer performs better in generating local high-frequency details, while the occlusion maps demonstrate that ConvTransformer has superiority on accurate pixel intensity prediction.}
  \label{Fig_1}
  \smallskip
  }
\makeatother

\maketitle
\begin{abstract}
    Deep Convolutional Neural Networks (CNNs) are powerful models that have achieved excellent performance on difficult computer vision tasks. Although CNNs perform well whenever large labeled training samples are available, they work badly on video frame synthesis due to objects deforming and moving, scene lighting changes, and cameras moving in video sequence. In this paper, we present a novel and general end-to-end architecture, called \textbf{convolutional Transformer} or \textbf{ConvTransformer}, for video frame sequence learning and video frame synthesis. The core ingredient of ConvTransformer is the proposed attention layer, i.e., multi-head convolutional self-attention layer, that learns the sequential dependence of video sequence. ConvTransformer uses an encoder, built upon multi-head convolutional self-attention layer, to encode the sequential dependence between the input frames, and then a decoder decodes the long-term dependence between the target synthesized frames and the input frames. Experiments on video future frame extrapolation task show ConvTransformer to be superior in quality while being more parallelizable to recent approaches built upon convolutional LSTM (ConvLSTM). To the best of our knowledge, this is the first time that ConvTransformer architecture is proposed and applied to video frame synthesis.
\end{abstract}

\section{Introduction}
    Video frame synthesis, comprising interpolation and extrapolation two subtasks, is one of the classical and fundamental problems in video processing and computer vision community. The abrupt motion artifacts and temporal aliasing in video sequence can be suppressed with the help of video frame synthesis, and hence it can be applied to numerous applications ranging from motion deblurring \cite{TimBrooks}, video frame rate up-sampling \cite{R.Castagno, WenboBao}, video editing \cite{WenqiRen, Zitnick}, novel view synthesis \cite{JohnFlynn} to autonomous driving \cite{HuazheXu}.

    Numerous solutions, over the past decades, have been proposed for this problem, and achieved substantial improvement. Traditional video frame synthesis pipeline usually involves two consecutive steps, i.e., optical flow estimation and optical based frame warping \cite{werlberger2011optical, yu2013multi}. The quality of synthesized frames, however, will be easily affected by the estimated optical flow. Recent advancements in deep neural networks have successfully improved a number of tasks, such as classification \cite{Krizhevsky}, segmentation \cite{Ronneberger} and visual tracking \cite{BoLi}, and also promoted the development of video frame interpolation and extrapolation.

    Long \emph{et al.} treated the video frame interpolation as an image generating task, and then trained a generic convolutional neural network (CNN) to directly interpolate the in-between frame \cite{2016Long}. However, due to the difficulty for generic CNN to capture the multi-modal distribution of video frames, severe blurriness exists in their interpolated results. Lately, Niklaus \emph{et al.} considered frame interpolation as a local convolution problem, and proposed adaptive convolutional operation \cite{SimonNiklaus} and separable convolutional process \cite{SimonNiklaus2} for each pixel in the target interpolated frame. However, these kernel-based algorithms typically suffer from heavy computation. Instead of only relying on CNN, the optical flow embedded neural netoworks are also investigated and applied to interpolated middle frames. The deep voxel flow (DVF) method, for instance, implicitly incorporate 3D optical flow across the time and space to synthesize middle frames. Besides, Bao \emph{et al.} proposed an explicitly optical flow information embedded algorithm, namely DAIN \cite{bao2019depth}, in which the depth map \cite{Chen, Zhengqi} and optical flow \cite{Zhengqi} are explicitly generated by the pretrained sub-networks \cite{Zhengqi, Chen, Zhengqi}, and is used to guide the contextual features capturing pipeline. Although these methods could interpolate perceptually well frames with the accurately estimated flow generated by the pretrained optical flow estimation sub-networks \cite{Zhengqi, Chen, Zhengqi}, the optical flow estimation networks are sensitive to the artificial marked dataset. Besides, these interpolation methods are mainly developed on two consecutive frames, while the high-order motion information of video frame sequence is ignored, and not well exploited. Furthermore, these specially designed interpolation methods could not be applied to another video frame synthesis task, i.e. video frame extrapolation, because the latter frame used to estimate flow can not be provided in extrapolation task.


    Unlike the video frame interpolation task, video frame extrapolation can be treated as a conditional prediction problem, that is, the future frames are predicted by the previous video frame sequence in a recurrent model. As a pioneer of recurrent-based extrapolation method, Shi \emph{et al.} proposed a convolutional LSTM (ConvLSTM) \cite{shi2015convolutional} to recurrently generate future frames. However, due to the capability limitation of this simple and generic recurrent architecture, the predicted frames usually suffer from blurriness. With the advances in research, several ConvLSTM variations \cite{villegas2017decomposing, William, 2019Inception} were proposed to improve the performance. Specifically, Villegas \emph{et al.} investigated a motion and content decomposition LSTM model, i.e. MCNet \cite{villegas2017decomposing}, to predict future frames. PredNet \cite{William} uses the top-down context to guide the training process. Besides, the inception operation is introduced in LSTM to obtain a broadly view for synthesizing \cite{2019Inception}. While the extrapolated frames look somewhat better, the accurate pixel estimation is still challenging for multi-modal nature scenes. Additionally, the recurrent model of ConvLSTM and its variations are hard to train, and meanwhile suffer from heavy computational burden when the length of sequence is large.

    Aside from these recurrent models, the plain CNN architecture based methods are also investigated to generate future frames. Liu \emph{et al.} developed a deep voxel flow (DVF) incorporated neural network to generate the future frames \cite{liu2017video}, in which the 3D optical flow across time and space are implicitly estimated. However, due to the lack of multi-frames joint training, DVF \cite{liu2017video} has a limitation in long-term multiple frames prediction. Different from these implicitly motion estimation methods, Wu \emph{et al.} proposed an explicitly motion detection and semantic estimation method to extrapolate future frames \cite{Wu}. Although it performs well in situations where there are explicitly foreground motions, it suffers from a restriction in nature video scenes, where the distinction between foreground motion target and background is not clear.

    Although these unidirectional prediction methods, i.e., along the frame sequence, can be applied to video frame interpolation task, these methods will suffer from heavy performance degradation, as compared with state-of-the-art interpolation algorithms. It is because the bi-directional information provided by latter frames can not be exploited to guide the middle frames generating pipeline in these recurrent and forward CNN based extrapolation models.


    In order to bridge this gap, we propose, in this work, a general video frame synthesis architecture, named convolutional Transformer (ConvTransformer), which unifies and simplifies the video frame extrapolation and video frame interpolation pipeline as an end-to-end encoder and decoder problem. A multi-head convolutional self-attention is proposed to model the long-range dependence between the frames in video sequence. As compared with previously elaborately designed interpolation methods, ConvTransformer is a simple, but efficient architecture in extracting the high-order motion information existing in video frame sequence. Besides, ConvTransformer, in comparison with previous ConvLSTM based recurrent models \cite{shi2015convolutional, villegas2017decomposing, William, 2019Inception}, ConvTransformer can be implemented in parallel both in training and testing stage.

    We evaluate ConvTransformer on several benchmarks, i.e. UCF101 \cite{soomro2012ucf101}, Vimeo90K \cite{xue2019video}, Sintel \cite{janai2017slow}, REDS \cite{Nah_2019_CVPR_Workshops_REDS}, HMDB \cite{Kuehne11} and Adobe240fps \cite{su2017deep}. Experimental results demonstrate that ConvTransformer performs better in extrapolating future frames when compared with previous ConvLSTM based recurrent models, and achives competitive results against state-of-the-art elaborately designed interpolation algorithms.

    The main contributions of this paper are therefore as follows.
    \begin{itemize}
        \item A novel architecture named ConvTransformer is proposed for video frame synthesis.
        \item A novel attention assigned multi-head convolutional self-attention is proposed for modeling the long-range spatial and temporal dependence on video frame sequence.
        \item The effectiveness and superiority of the proposed ConvTransformer have been comprehensively analyzed in this paper.
    \end{itemize}

    \begin{figure*}[] \centering
        \centering
        \label{Fig_2}
    	\includegraphics[width=17cm]{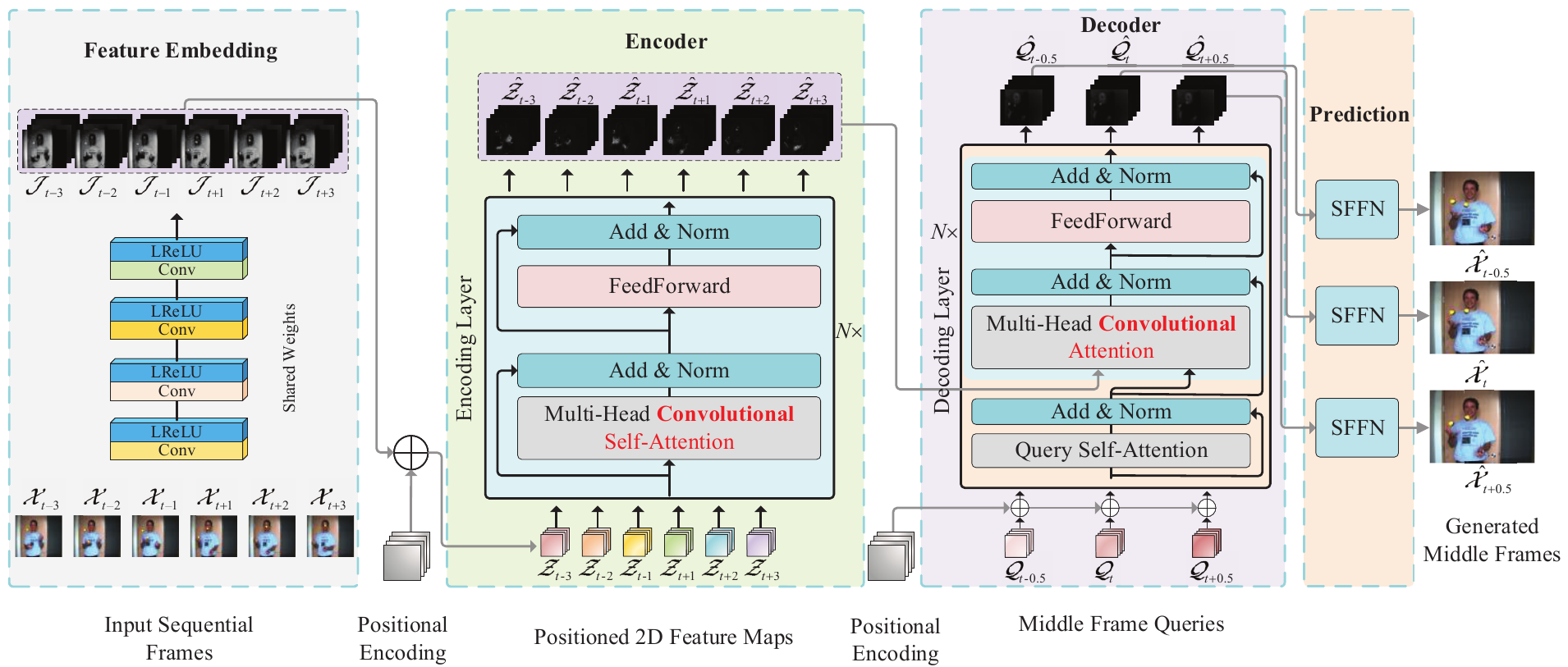}
    	\caption{An overview of ConvTransformer ${\mathcal G}_{\theta_{G}}$. ConvTransformer ${\mathcal G}_{\theta_{G}}$ consists of four parts: feature embedding, encoder, decoder and prediction. In feature embedding part, a shared convolutional neural network transforms the input video frame image in RGB format to a compact feature maps space. In the following encoder step, several stacked encoding layers incorporating multi-head convolutional self-attention module are used to extract the long-term dependence among the input video sequence. Thirdly, the encoded video feature maps sequence and target query frames are passed to the decoder layers. The sequential dependence between target frames and input video sequence is decoded in this step. Finally, the target frame sequence in RGB format is generated in the prediction step with the shared network SFFN.}
    \end{figure*}
\section{Related Work}
\subsection{Video Frame Synthesis}
    Video frame synthesis, including interpolation and extrapolation two subtasks, is a hot research topic and has been extensively investigated in the literature \cite{long2016learning, liu2017video, vondrick2016generating, SimonNiklaus, SimonNiklaus2, villegas2019higFidelity, villegas2017decomposing, villegas2017learning}. In the following, we give a detailed discussion on video frame interpolation and extrapolation, respectively.

    \textbf{Video frame interpolation.} To interpolate the middle frames between two adjacent frames, traditional approaches mainly consist of two steps: motion estimation, and pixel synthesis. With the accurately estimated motion vectors or optical flow, the target frames can be interpolated via the conventional warping method. However, accurate motion estimation still remains a challenging problem because the motion blur, occlusion and brightness change frequently appears in nature images.

    The recent years have witnessed significant advances in deep neural networks which have shown great power in solving a variety of learning problems \cite{Krizhevsky, Ronneberger, BoLi, WenqiRen, Zitnick}, and have attracted much attention on applying it to video frame interpolation task. Long \emph{et al.} developed a general CNN to directly synthesize the middle frames \cite{long2016learning}. However, a generic CNN, only stacking several convolutional layers, is limited in capturing the dynamic motion of nature scenes, and thereby their method usually yields severe visual artifact, e.g. motion blur and ringing artifact. Lately, Niklaus \emph{et al.} treated the frame interpolation problem as a local convolutional kernel prediction problem, and proposed adaptive convolution operation \cite{SimonNiklaus} and separable convolution process \cite{SimonNiklaus2} for each pixel in frames. To some extent, these methods perform better than previous general CNN based method. But even so, high memory footprint is the typical characteristic of these kernel-based algorithms, and thereby restricts its application. Unlike these CNN and kernel-based algorithms, in literature \cite{Meyer}, a pixel-phase information learning network PhaseNet \cite{Meyer} was proposed for frame interpolation. To a certain extent, it suppresses the artifacts as compared with previous methods. The performance, however, will be easily affected by complicated factors, e.g. large motion and disparity. With the rapid development of exploiting deep neural networks for optical flow estimation \cite{Zhengqi} and depth map generating \cite{Chen, Zhengqi}, several implicitly or explicitly flow and depth map guided neural networks \cite{jiang2018super, bao2019memc, park2020bmbc, bao2019depth} have been investigated for interpolating middle frames. The typically pipeline of these algorithms lies in optical flow or depth information estimating realised by pre-trained sub-networks \cite{Zhengqi, Chen, Zhengqi} firstly, and then a warping layer is introduced to adaptively warping the input frames subsequently, and finally a information fusing layer is built to generate the target middle frames. These methods work well on anticipating occlusion and artifacts, and thus interpolate sharp images. However, these methods are restricted by the pre-trained optical flow sub-network PWC-Net \cite{Zhengqi} and depth map estimation sub-network \cite{Chen, Zhengqi} which can be easily affected by the training set. Last but not least, the architectures of these methods are all specially and elaborately designed, and hence the generalization to another video frame synthesis task, i.e. video frame extrapolation, is limited.

    \textbf{Video frame extrapolation.} Extrapolating future video frames in video content still remains a challenging task because of the multi-model of nature videos and the unexpected incident in the future. Traditional solutions take this problem as a recurrent prediction problem \cite{shi2015convolutional, villegas2017decomposing, William}. Shi \emph{et al.} proposed a convolutional LSTM (ConvLSTM) architecture \cite{shi2015convolutional}, a pioneer recurrent model, to generate future frames. Given to the flexibility of nature scenes and limited representation ability of this simple architecture, the predicted frames naturally suffer from blurriness. Lately, several improved algorithms based on ConvLSTM architecture were proposed, and, to some extent, achieved better performance. Specifically, Villegas \emph{et al.} proposed a decompostion LSTM model, namely MCNet \cite{villegas2017decomposing}, in which the motion and content are decomposed respectively. In order to utilize the contextual information, Lotter \emph{et al.} proposed a top-down context guided LSTM model PredNet \cite{William}. Besides, the inception mechanism is introduced in LSTM to obtain a broadly view for synthesizing \cite{2019Inception}. Aside from these LSTM based recurrent models, the plain CNN architectures based models are also investigated to generate future frames. Liu \emph{et al.} proposed a 3D optical flow, across the time and space, guided neural networks, dubbed DVF \cite{liu2017video}, to extrapolated frames. However, due to the limitation in multi-frames joint training, DVF \cite{liu2017video} cannot work well on long-term multiple frames prediction. Apart from these implicitly motion estimation and utilising methods, Wu \emph{et al.} proposed an explicitly motion detection and semantic estimation method to extrapolate future frames \cite{Wu}. Although it works well when there are explicitly foreground motions in frame sequence, e.g. a car and a runner, it suffers from a restriction in nature video scenes, where the distinction between the forground and background is not clear.

    Although recent years have witness a great progress in video interpolation and extrapolation two sub-tasks, there is still a lack of a general and unified video frame synthesis architecture that can perform well on both of these two sub-tasks. In order to overcome this issue, we propose an unified architecture, named convolutional Transformer (ConvTransformer), for video frame synthesis. A simple but efficient multi-head convolutional self-attention architecture is proposed to model the long-range dependence between the video frame sequence. The experiment results conducted on several benchmarks typically demonstrate that the proposed ConvTransformer works well both in video frame interpolation and extrapolation. To the best of our knowledge, it is the first time that ConvTransformer architecture is proposed, and has been successfully applied to video frame synthesis.

\subsection{Transformer Network}
    Transformer \cite{vaswani2017attention} is a novel architecture for learning long-range sequential dependence, which abandons the traditional building style of directly using RNN or LSTM architecture. It has been successfully applied to numerous natural language processing (NLP) tasks, such as machine translation \cite{kenton2019bert} and speech processing \cite{dong2018speech, wang2020transformer}. Recently, the basic Transformer architecture has been successfully applied to the field of image generation \cite{parmar2018image}, image recognition \cite{dosovitskiy2020image}, and object detection \cite{carion2020end}. Specifically, Nicolas \emph{et al.} proposed the DETR \cite{carion2020end} object detection method, and it achieves competitive result on COCO dataset as compared with Faster R-CNN \cite{renNIPS15fasterrcnn}. Through collapsing the spatial dimensions (two dimensions) into one dimension, DETR reasons about the relations of pixels and the global image context. Although DETR has successfully applied Transformer for computer vision task object detection, it is hard to use the basic Transformer to model the long term dependence among the two dimensional video frames, which are not only temporally related, but also spatially related. In order to overcome this issue, a convolutional Transformer (ConvTransformer) is proposed in this work, and has been successfully applied to video frame synthesis including interpolation and extrapolation two subtasks.
\section{Convolutional Transformer Architecture}
    The overall architecture of ConvTransformer ${\mathcal{G}}_{\theta_{G}}$, as shown in Figure 2, consists of five main components, that is, feature embedding module ${\mathcal F}_{\theta_{F}}$, positional encoding module ${\mathcal P}_{\theta_{P}}$, encoder module ${\mathcal E}_{\theta_{E}}$, decoder module ${\mathcal D}_{\theta_{D}}$, and synthesis feed-forward network ${\mathcal S}_{\theta_{S}}$. In this section, we first provide an overview of video frame synthesis pipeline realised by ConvTransformer architecture, and then make an illustrated introduction of the proposed ConvTransformer. Finally, the implementation details and training loss are introduced.
    \begin{figure*}[]
        \centering
    	\includegraphics[width=.9\linewidth]{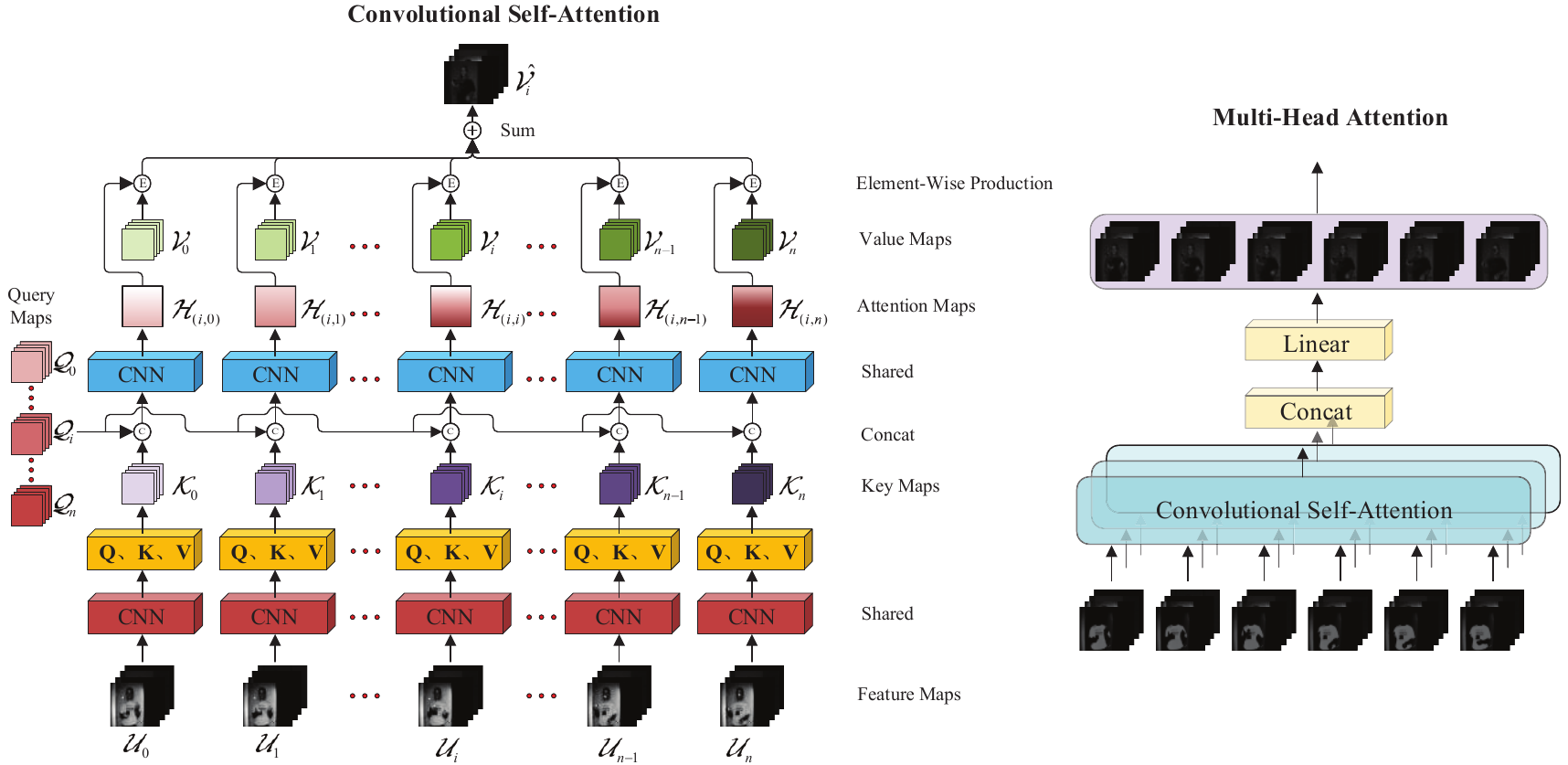}
    	\caption{(left) Convolutional Self-Attention. (right) Multi-Head Attention in parallel.}
    \end{figure*}
\subsection{Algorithm Review}
    Given a video frame sequence ${\mathcal{\tilde X}}{\rm{ = \{ }}{{\mathcal{\tilde X}}_{0}}{\rm{, }}{{\mathcal{\tilde X}}_{1}}{\rm{, }}\cdot\cdot\cdot{\rm{ ,}}{{\mathcal{\tilde X}}_{n}}{\rm{ \} }} \in {\mathbb{R}^{{{H \times W \times C}}}}$, where $n$ is the length of sequence and $H$, $W$ and $C$ denote height, width, and the number of channels, respectively, our goal is to synthesize intermediate frames ${\mathcal{\hat X}}= \{{{\mathcal{\hat X}}_{i+t_0}}, {{\mathcal{\hat X}}_{i+t_1}},\cdot\cdot\cdot, {{\mathcal{\hat X}}_{i+t_k}} \}$ at time $t_m \in [0,1]$, or extrapolate future frames ${\mathcal{\hat X}}= \{{{\mathcal{\hat X}}_{n+1}}, {{\mathcal{\hat X}}_{n+2}},\cdot\cdot\cdot, {{\mathcal{\hat X}}_{n+m_k}} \}$ at order ${{m}_k}\in \mathbb{N}$. Specifically, the future frames forecasting problem can be viewed as:
    \begin{equation}\label{Eq3}
            {\hat{\mathcal X}}_{n+1},\cdot\cdot\cdot,{\hat{\mathcal X}}_{n+m_k} = {\mathop {\rm argmax}\limits_{{{\mathcal X}}_{n+1},\cdot\cdot\cdot,{{\mathcal X}}_{n+m_k}}}{P}({{\mathcal X}}_{n+1},\cdot\cdot\cdot,{{\mathcal X}}_{n+m_k}|{{\mathcal{\tilde X}}})
    \end{equation}
    Here, $P(\cdot|\cdot)$ stands for conditional probability operation.

    Firstly, the \textbf{Feature Embedding} module embeds the input video frames, and then generates representative feature maps. Subsequently, the extracted feature maps of each frame are added with the positional maps, which are used for positional identity. Next, the positioned frame feature maps are passed as inputs to the \textbf{Encoder} to exploit the long-range sequential dependence among each frame in video sequence. After getting the encoded high-level feature maps, the high-level feature maps and positioned frame queries are simultaneously passed into the \textbf{Decoder}, and then the sequential dependence between the query frames and input sequential video frames is decoded. Finally, the decoded feature maps are fed into the \textbf{Synthesis Feed-Forward Networks (SFFN)} to generate the final middle interpolated frames or extrapolated frames.
\subsection{Feature Embedding ${\mathcal F}_{\theta_{F}}$}
    In order to extract a compact feature representation for subsequent effective learning, a representative feature is computed by a 4-layer convolution with Leaky ReLu activation function and hidden dimension $d_{model}$. Given a video frame ${\mathcal{X}}_{i} \in {\mathbb{R}^{{{H \times W \times 3}}}}$, the embedded feature maps ${\mathcal{H}}_{i} \in {\mathbb{R}^{{{H \times W \times d_{model}}}}}$ can be presented with the following equation:
    \begin{align}\label{Eq3}
        {\mathcal{J}}_{i} = {\mathcal F}_{\theta_{F}}({\mathcal{X}}_{i}), \quad {i \in[1, n]}
    \end{align}

    It is worth mentioning that all input video frames share not only the same embedding net architecture ${\mathcal F}_{\theta_{F}}$, but also the parameters ${\theta_{F}}$.

\subsection{Positional Encoding ${\mathcal P}_{\theta_{P}}$}
    Since our model contains no recurrence across the video frame sequence, some information about relative or absolute position of the video frames must be injected in the frames’ feature maps, so that the order information of the video frame sequence can be utilized. To this end, ``positional encodings" are added at each layer in encoder and decoder. It is noted that the positional encoding in ConvTransformer is a 3D tensor which is different from that in original Transformer architecture built for vector sequence. The positional encoding has the same dimension as the frame feature maps, so that they can be summed directly. In this work, we use sine and cosine functions of different frequencies to encode the position of each frame in video sequence:
        \begin{align}\label{Eq3}
        {\rm{Pos\_Ma}}{{\rm{p}}_{(p,(i,j,2k))}} = \sin (p/{10000^{2k/{d_{model}}}})
        \end{align}
        \begin{align}\label{Eq4}
        {\rm{Pos\_Ma}}{{\rm{p}}_{(p,(i,j,2k+1))}} = \cos (p/{10000^{2k/{d_{model}}}})
        \end{align}
    where $p\in[1, n]$ is the positional token, $(i, j)$ represents the spatial location of features and the channel dimension is noted as $2k$. That is, each dimension of the positional encoding corresponds to a sinusoid. The wavelengths form a geometric progression from $2\pi$ to $10000*2\pi$. We choose this function because it would allow the model to easily learn to attend relative positions for any fixed offset $m$, ${\rm{Pos\_Ma}}{{\rm{p}}_{(p+m)}}$ can be represented as a linear function of ${\rm{Pos\_Ma}}{{\rm{p}}_{(p)}}$.

    Given an embedded feature maps ${\mathcal{J}}_{i}$, the positioned embedding process can be viewed as the following equation:
        \begin{align}\label{Eq5}
            {\mathcal{Z}}_{i} = {\mathcal{J}}_{i} \oplus  {\rm{Pos\_Ma}}{{\rm{p}}_{(i)}}, \quad {i \in[1, n]}
        \end{align}
    where the $\oplus$ operation represents element-wise tensor addition.

\subsection{Encoder ${\mathcal E}_{\theta_{E}}$ and Decoder ${\mathcal D}_{\theta_{D}}$}
    \textbf{Encoder:} As shown in Figure 2, the encoder is modeled as a stack of $N$ identical layers consisting of two sub-layers, i.e., multi-head convolutional self-attention layer and a simple 2D convolutional feed-forward network. The residual connection is adopted around each of the two sub-layers, followed by group normalization \cite{wu2018groupnormalization}. To facilitate these residual connections, all sub-layers in the model, as well as the embedding layers, produce outputs of the same dimensional $d_{model}$. Given a positioned feature sequence ${\mathcal{Z}}{\rm{ = \{ }}{{\mathcal{Z}}_{0}}{\rm{, }}{{\mathcal{Z}}_{1}}{\rm{, }}\cdot\cdot\cdot{\rm{ ,}}{{\mathcal{Z}}_{i}}{\rm{, }}\cdot\cdot\cdot{\rm{, }}{{\mathcal{Z}}_{n-1}}{\rm{, }}{{\mathcal{Z}}_{n}}{\rm{ \} }} \in {\mathbb{R}^{{{H \times W \times d_{model}}}}}$, the identified feature sequence ${\mathcal{\hat Z}}{\rm{ = \{ }}{{\mathcal{\hat Z}}_{0}}{\rm{, }}{{\mathcal{\hat Z}}_{1}}{\rm{, }}\cdot\cdot\cdot{\rm{, }}{{\mathcal{Z}}_{n}}{\rm{ \} }} \in {\mathbb{R}^{{{H \times W \times d_{model}}}}}$ can be learned, and the encoding operation can be represented as:
        \begin{align}\label{Eq3}
            {\mathcal{\hat Z}} = {\mathcal{E}}_{\theta_{E}}({\mathcal{Z}})
        \end{align}
    \textbf{Decoder:} The decoder is also composed of a stack of $N$ identical layers, which consists of three sub-layers. In addition to the two sub-layers as implemented in \textbf{Encoder}, an additional layer called query self-attention is inserted to perform the convolutional self-attention over the output frame queries. Given a query sequence ${\mathcal{Q}}{\rm{=\{ }}{{\mathcal{Q}}_{0}}{\rm{, }}{{\mathcal{Q}}_{1}}{\rm{, }}\cdot\cdot\cdot{\rm{, }}{{\mathcal{Q}}_{n}}{\rm{ \} }} \in {\mathbb{R}^{{{H \times W \times d}}}}$, the decoding process can be conducted as:
        \begin{align}\label{Eq3}
            {\mathcal{\hat Q}} = {\mathcal{D}}_{\theta_{D}}({\mathcal{\hat Z}},{\mathcal{Q}})
        \end{align}

    It should be emphasized that the encoding and decoding process are all conducted in parallel.

\subsection{Multi-Head Convolutional Self-Attention}
    We call our particular attention ``Convolutional Self-Attention" ( as shown in Figure 3), which is computed upon feature maps. The convolutional self-attention operation can be described as mapping a query map and a set of key-value map pairs to an output, where the query map, key maps, value maps, and output are all 3D tensors.
    Given an input comprised of sequential feature maps ${\mathcal{U}}{\rm{ = \{ }}{{\mathcal{U}}_{0}}{\rm{, }}{{\mathcal{U}}_{1}}{\rm{, }}\cdot\cdot\cdot{\rm{, }}{{\mathcal{U}}_{n}}{\rm{ \} }} \in {\mathbb{R}^{{{H \times W \times d_{model}}}}}$, we apply convolutional sub-network to generate the query map and paired key-value map of each frame, i.e., ${\mathcal{U'}}{\rm{ = \{ }}{ \{ {\mathcal{Q}}_{0},{\mathcal{K}}_{0},{\mathcal{V}}_{0}\}}{\rm{, }}{\{ {\mathcal{Q}}_{1},{\mathcal{K}}_{1},{\mathcal{V}}_{1}\}}{\rm{, }}\cdot\cdot\cdot{\rm{, }}{ \{ {\mathcal{Q}}_{n},{\mathcal{K}}_{n},{\mathcal{V}}_{n}\}}{\rm{ \} }} \in {\mathbb{R}^{{{H \times W \times d_{model}}}}}$.

    Given a set of ${\{ {\mathcal{Q}}_{i},{\mathcal{K}}_{i},{\mathcal{V}}_{i}\}}$ of frame ${{\mathcal{U}}_{i}}$, the attention map ${{\mathcal H}_{(i,j)}} \in {\mathbb{R}^{{{H \times W \times 1}}}}$ of frame ${{\mathcal{U}}_{i}}$ and ${{\mathcal{U}}_{j}}$ can be generated by applying a compatible sub-network ${\mathcal{M}}_{\theta_{M}}$ to the query map ${\mathcal{Q}}_{i}$ with the corresponding key map ${\mathcal{K}}_{j}$, which can be represented as following equation:
        \begin{align}\label{Eq3}
            {{\mathcal H}_{(i,j)}} = {\mathcal{M}}_{\theta_{M}}({\mathcal{Q}}_{i}, {\mathcal{K}}_{j})
        \end{align}

    After getting all the corresponding attention map ${{\mathcal H}_{(i)}}=\{{{\mathcal H}_{(i,1)}}, {{\mathcal H}_{(i,2)}},\cdot\cdot\cdot,{{\mathcal H}_{(i,n)}}\} \in {\mathbb{R}^{{{H \times W \times 1}}}}$, we make a concatenation operation of ${{\mathcal H}_{(i)}}$ in the third dimension, and then a SoftMax operation is applied to ${{\mathcal H}_{(i)}} \in {\mathbb{R}^{{{H \times W \times n}}}}$ along the dimension $dim=3$.
    \begin{align}\label{Eq3}
        {{\mathcal H}_{(i)}} = SoftMax({{\mathcal H}_{(i)}})_{dim}, \quad dim=3
    \end{align}

    Finally, the output ${{\hat {\mathcal V}}_i}$ can be obtained with summation of the element wise production with attention map ${{\mathcal H}_{(i,j)}}$ and the corresponding value map ${{\mathcal V}_{j}}$. This operation can be represented as:
    \begin{align}\label{Eq3}
        {{\mathcal V}_{(i)}} = \sum\limits_{j = 1}^n {{{\mathcal H}_{(i,j)}}{{\mathcal V}_j}}
    \end{align}

    In order to jointly attend to information from different representation subspaces at different feature spaces, a multi-head pipeline is adopted. The process can be viewed as:
    \begin{align}\label{Eq3}
        {\rm MultiHead}({{\hat {\mathcal V}}_{i}}) = {\rm Concat}({{\mathcal{\hat V}}_{i_1}},\cdot\cdot\cdot,{{\mathcal{\hat V}}_{i_h}})
    \end{align}

\subsection{Synthesis Feed-Forward Network ${\mathcal S}_{\theta_{S}}$}
    In order to synthesize the final photo realistic video frames, we construct a frame synthesis feed-forward network, which consists of 2 cascaded sub-networks built upon a U-Net-like structure. The frames states $\mathcal{\hat Q}$ decoded from previous decoder are fed into SFFN in parallel. This process can be represented as:
        \begin{align}\label{Eq3}
            {{\mathcal{\hat X}}}_{i} = {\mathcal S}_{\theta_{S}}({\mathcal{\hat Q}}_{i}), \quad {i \in[1, N']}
        \end{align}

\subsection{Initialization of Query Set $\mathcal{Q}$}
    As an indispensable part of \textbf{Decoder}, query set $\mathcal{Q}$ is critical for accurate extrapolation and interpolation. Specifically, given 4 input frames $\mathcal{X}=\{\mathcal{X}_1,\mathcal{X}_2,\mathcal{X}_3,\mathcal{X}_4\}$, ConvTransformer extrapolates 3 frames $\mathcal{\hat X}=\{\mathcal{ \hat X}_1,\mathcal{ \hat X}_2,\mathcal{ \hat X}_3\}$. The query $\mathcal{Q}_i$ equals to the embedded feature maps of the last input frame, i.e., $\mathcal{J}_4$. On the other hand, given a 6-frame sequence $\mathcal{X}=\{\mathcal{X}_{t-3}, \mathcal{X}_{t-2}, \mathcal{X}_{t-1}, \mathcal{X}_{t+1}, \mathcal{X}_{t+2}, \mathcal{X}_{t+3}   \}$, ConvTransormer interpolates 3 frames between frame $\mathcal{X}_{t-1}$ and $\mathcal{X}_{t+1}$, i.e., $\mathcal{\hat X}=\{\mathcal{ \hat X}_{t-0.5},\mathcal{ \hat X}_t,\mathcal{ \hat X}_{t+0.5}\}$. The query $\mathcal{Q}_i$ is calculated by the element-wise average calculation of two adjacent frames' feature maps, i.e., $\mathcal{J}_{t-1}$ and $\mathcal{J}_{t+1}$. Specifically, $\mathcal{Q}_i= \mathcal{J}_{t-1} \Delta \mathcal{J}_{t+1}$, where the operation $\Delta$ is element-wise average calculation.

\subsection{Training Loss ${{\mathcal L}_{{{\mathcal G}_{{\theta _G}}}}}$}
    In this work, we choose the most widely used content loss, i.e., pixel-wise \textbf{MSE loss}, to guide the optimizing process of ConvTransformer. The \textbf{MSE loss}
    is calculated as:
        \begin{align}\label{Eq4}
            {{\mathcal L}_{{{\mathcal G}_{{\theta _G}}}}} = \frac{1}{N'}\sum\limits_{i = 1}^{N'} {\mathop {\left\| {{\mathcal{\hat{X}}_{i}} - {\mathcal Y}_i} \right\|}\nolimits_2 }
        \end{align}
    Here, $N’$ stands for the number of synthesized results, $\mathcal{\hat X}_i$ represents the synthesized target frame, and the $\mathcal{Y}_i$ is the corresponding groundtruth.

\section{Experiments and Analysis}
    In this section, we provide the details for experiments, and results that demonstrate the performance and efficiency of ConvTransformer, and compare it with previous proposed specialized video frame extrapolation methods, elaborately designed video frame interpolation algorithms and general video frame synthesis solutions on several benchmarks. Besides, to further validate the proposed ConvTransformer, we conducted several ablation studies.

    \begin{table}[!t]
      \centering
      \caption{Details about trainset, validationset and testset}
      \resizebox{0.95\linewidth}{!}{
        \begin{tabular}{ccc}
          \hline \hline
          Training & Validation  & Test \\
          \hline
          \multirow{5}{*}{\begin{tabular}[c]{@{}c@{}}Vimeo90K\\ ( 64612 sequences) \\ Adobe240fps \\ (2120 sequences)\end{tabular}} & \multirow{5}{*}{\begin{tabular}[c]{@{}c@{}}Vimeo90K\\ ( 20 sequences) \\ Adobe240fps \\ (120 sequences)\end{tabular}} & \makecell[l]{Viemo90K (935 sequences)} \\
          & & \makecell[l]{UCF101 (2533 sequences)}\\
          & & \makecell[l]{Adobe240fps (2660 sequences)}\\
          & & \makecell[l]{Sintel (1581 sequences)}\\
          & & \makecell[l]{HMDB (2684 sequences)}\\
          \hline\hline
        \end{tabular}
      }
    \end{table}

\subsection{Datasets}
    To create the trainset of video frame sequence, we leverage the frame sequence from the Vimeo90K \cite{xue2019video} and Adobe240fps \cite{su2017deep} dataset. On the other hand, we also exploit several other widely used benchmarks, including UCF101 \cite{2012UCF101}, Sintel \cite{janai2017slow}, REDS \cite{Nah_2019_CVPR_Workshops_REDS} and HMDB \cite{Kuehne11}, for testing. Table 1 represents the details about training, validation and testing sets.
    \begin{figure*}[]
        \centering
        \label{Fig_3}
    	\includegraphics[width=0.99\linewidth]{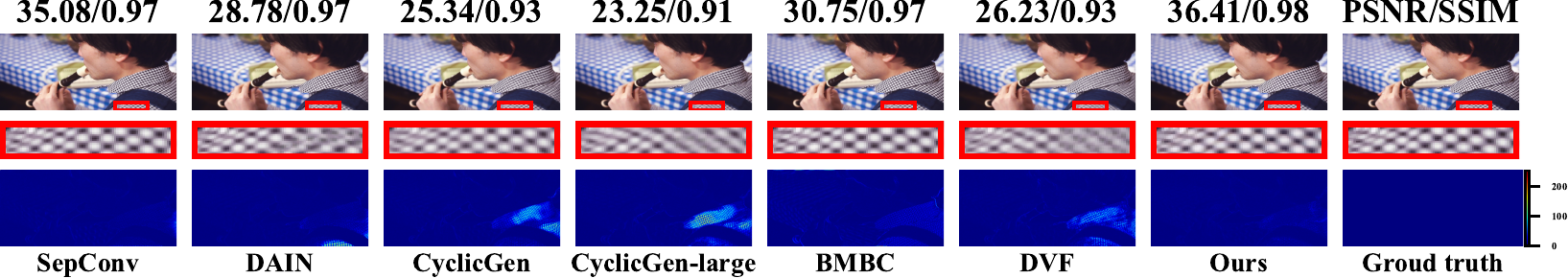}
    	\caption{Visual Comparisons of ConvTransformer with other state-of-the-art video frame interpolation methods:
                 SepConv \cite{SimonNiklaus2}, DAIN \cite{bao2019depth}, CyclicGen \cite{liu2019cyclicgen}, CyclicGen-large \cite{liu2019cyclicgen}, BMBC \cite{park2020bmbc} and DVF \cite{liu2017video}.}
    \end{figure*}
    \begin{table*}[!ht]
    	\renewcommand{\arraystretch}{0.9}
    	\caption{Video frame extrapolation: Quantitative evaluation of ConvTransformer with state-of-the-art methods.}
    	\centering
    	\resizebox{0.99\textwidth}{!}{
    		\begin{tabular}{c c c c c c c c c c c c c c c c c c}
    			\hline\hline \\[-4mm]
                 \multirow{3}{*}{Model} & \multicolumn{8}{c}{Next frame} & & \multicolumn{8}{c}{Next 3 frames}\\
                \cline{2-9}\cline{11-18}
                 & \multicolumn{2}{c}{UCF101 \cite{2012UCF101}} & \multicolumn{2}{c}{Adobe240fps \cite{su2017deep}} & \multicolumn{2}{c}{Vimeo90K  \cite{xue2019video}} &\multicolumn{2}{c}{Average} & & \multicolumn{2}{c}{UCF101 \cite{2012UCF101}} & \multicolumn{2}{c}{Adobe240fps \cite{su2017deep}} & \multicolumn{2}{c}{Vimeo90K \cite{xue2019video}} & \multicolumn{2}{c}{Average}\\
                 \cline{2-9}\cline{11-18}
                 & \makecell[c]{PSNR}
                 & \makecell[c]{SSIM}
                 & \makecell[c]{PSNR}
                 & \makecell[c]{SSIM}
                 & \makecell[c]{PSNR}
                 & \makecell[c]{SSIM}
                 & \makecell[c]{PSNR}
                 & \makecell[c]{SSIM}
                 &
                 & \makecell[c]{PSNR}
                 & \makecell[c]{SSIM}
                 & \makecell[c]{PSNR}
                 & \makecell[c]{SSIM}
                 & \makecell[c]{PSNR}
                 & \makecell[c]{SSIM}
                 & \makecell[c]{PSNR}
                 & \makecell[c]{SSIM}\\
                \hline

                \makecell[c]{DVF \cite{liu2017video}} & 29.1493  & 0.9181  & 28.7414 & 0.9254 & 27.8021 & 0.9073 &28.5642 & 0.9169 & & 26.1174 & 0.8779 & 25.8625 & 0.8598 & 24.5277 & 0.8432 &25.5025 &0.8603\\
                \cline{2-18}
                \makecell[c]{MCNet \cite{villegas2017decomposing}} & 27.6080 & 0.8504 & 28.2096 & 0.8796 & 28.6178 & 0.8726 & 28.1451 & 0.8675 & & 25.0179 & 0.7766 & 24.9485 & 0.7721 & 25.4455 & 0.7671 &25.1373 & 0.7719\\
                \cline{2-18}
                \makecell[c]{Ours} & \textbf{29.2814} & \textbf{0.9205} & \textbf{30.4233} & \textbf{0.9457} & \textbf{30.5161} & \textbf{0.9406} &\textbf{30.0736} &\textbf{0.9356} & & \textbf{26.7584} & \textbf{0.8874} & \textbf{28.0835} & \textbf{0.9045} & \textbf{27.1441} & \textbf{0.8926} & \textbf{27.3286} & \textbf{0.8948} \\
                \cline{2-18}
    			\hline\hline
    		\end{tabular}
    	}
    \end{table*}

\subsection{Training Details and Parameters Setting}
    Pytorch platform is used in our experiment for training. The experimental environment is the Linux operation system Ubuntu 16.04 LTS running on a sever with an AMD Ryzen 7 3800X CPU at 3.9GHz and a NVIDIA RTX 3090 GPU. In order to guarantee the convergence of ConvTransformer, the Adam optimizer is adopted for training, in which, the initial learning rate is set to $10^{-4}$ and is reduced with exponential decay, where the decay rate is 0.95 while the decaying quantity is 20000. The whole training proceeds for $6 \times {10^5}$ iterations. Besides, the length of input sequence is 4 for video frame prediction, while for the video interpolation task, the length of the input sequence is 6. The more setting details about ConvTransformer are represented in appendix file.

\subsection{Comparisons with state-of-the-arts}
    In order to evaluate the performance of the proposed ConvTransformer, we compare our finally trained ConvTransformer on several public benchmarks with state-of-the-art video frame synthesis method DVF \cite{liu2017video}, representative ConvLSTM based video frame extrapolation algorithm MCNet \cite{villegas2017decomposing}, and specially designed video frame interpolation solutions, i.e., SepConv \cite{SimonNiklaus2}, CyclicGen \cite{liu2019cyclicgen}, DAIN \cite{bao2019depth} and BMBC \cite{park2020bmbc}. For a fair comparison, we reimplemented and retrained these methods with the same trainset for training our ConvTransformer. Two widely used image quality metrics, i.e., peak signal to noise ratio (PSNR) \cite{Hor2010Image} and structural similarity (SSIM) \cite{Wang2004Image}, are adopted as the objective evaluation criteria. The quantitative results of video frame extrapolation are tabulated in Table 2, while Table 3 represents the quantitative comparison of video frame interpolation. Besides, the visual comparisons of synthesized images with zoomed details and residual occlusion maps are illustrated in Figure 1 and Figure 4, respectively.

    As observed in Table 2, the proposed ConvTransformer has given rise to better performance than DVF \cite{liu2017video} and MCNet \cite{villegas2017decomposing}. More concretely, taking the next frame extrapolation as an example, the relative performance gains of ConvTransformer over the DVF and MCNet \cite{villegas2017decomposing} models, in terms of index PSNR, are 2.7140dB and 1.8983dB on Vimeo 90k \cite{xue2019video}, 1.6819dB and 2.2137dB on Adeobe240fps \cite{su2017deep}, as well as 0.1321dB and 1.6734dB on UCF101 \cite{2012UCF101}. Besides, ConvTransformer, in terms of average comparison, achieves 1.5094dB and 1.9285dB advantage on DVF \cite{liu2017video} and MCNet \cite{villegas2017decomposing} respectively. Additionally, the superiority of ConvTransformer becomes larger in multiple future frames extrapolation. Specifically, ConvTransformer gains an advantage of 2.22dB in PSNR criterion over the method DVF \cite{liu2017video}, while it is 1.68dB in previous next frame extrapolation on the same benchmark Adobe240fps \cite{su2017deep}.

    Figure 1 visualizes the qualitative comparisons. As observed from the zoomed-in regions in Figure 1, the proposed ConvTransformer could extrapolate more photo-realistic frames, while the predicted frames generated from previous methods DVF \cite{liu2017video} and MCNet \cite{villegas2017decomposing} suffer from image degradation, such as image distortion and local smother. Besides, the residual occlusion maps suggest that ConvTransformer has superiority on accurate pixel value estimation, as compared with DVF \cite{liu2017video} and MCnet \cite{villegas2017decomposing}.

    In a nutshell, the quantitative and qualitative results prove that the proposed ConvTransformer, incorporating multi-head convolutional self-attention mechanism, can efficiently model the long-range sequential dependence in video frames, and then extrapolate high quality future frames.

    Except for the video frame extrapolation comparison, we also make a video frame interpolation comparison with the popular and state-of-the-art methods, including the general video frame synthesis method DVF \cite{liu2017video} and five specialized interpolation solutions, namely namely SepConv \cite{SimonNiklaus2}, CyclicGen \cite{liu2019cyclicgen}, CyclicGen-large \cite{liu2019cyclicgen}, DAIN \cite{bao2019depth} and BMBC \cite{park2020bmbc}. The performance indices of these methods are illustrated in Table 3. We can easily find that the ConvTransformer has attained better performance over the previous synthesis method DVF \cite{liu2017video}. In view of the PSNR and SSIM index, the proposed ConvTransformer has introduced a relative performance gain of 1.36dB and 0.0205 on average. On the other hand, as compared with specially designed interpolation methods, ConvTransformer outperforms most solutions, and achieves competitive results as compared with best algorithm DAIN \cite{bao2019depth}. Concretely, considering the PSNR and SSIM criteria, ConvTransformer achieves 31.57dB and 0.9151 on average, which is better than popular methods SepConv-Lf \cite{SimonNiklaus2}, CyclicGen \cite{liu2019cyclicgen}, CyclicGen-large \cite{liu2019cyclicgen} and BMBC \cite{park2020bmbc}. Although ConvTransformer does not outperform state-of-the-art interpolation method DAIN, the performance gap between them is not so large. Furthermore, as compared with elaborately designed algorithm DAIN \cite{bao2019depth}, ConvTransformer is a more general model, which not only works well on video frame interpolation, but also performs well on video frame extrapolation. On the contrary, the solution DAIN \cite{bao2019depth} could not be used for the video frame extrapolation task, because the latter frame , used for estimating and depth map and optical flow map, cannot be provided in video frame extrapolation task.

    According to the visual and quantitative comparisons above, two dominant conclusions can be drawn. First,  ConvTransformer is an efficient model for synthesizing photo-realistic extrapolation and interpolation frames. Second, in contrast to the specially designed extrapolation and interpolation methods, i.e.,  MCnet \cite{villegas2017decomposing}, SepConv \cite{SimonNiklaus2}, CyclicGen \cite{liu2019cyclicgen}, CyclicGen-large \cite{liu2019cyclicgen}, DAIN \cite{bao2019depth} and BMBC \cite{park2020bmbc}, ConvTransformer is a unify and general architecture, that is, it performs well in both two subtasks.

    \begin{table*}[!h]
    	\caption{Video frame interpolation: Quantitative evaluation of ConvTransformer with state-of-the-art methods.}
    	\centering
    	\label{table 1}
    	\resizebox{0.97\linewidth}{!}{
    		\begin{tabular}{c c c c c c c c c c c c c c c c }
    			\hline\hline \\[-4mm]
                 \multicolumn{2}{c}{\multirow{2}*{Model}}&   \multicolumn{2}{c}{Sintel \cite{janai2017slow}} & \multicolumn{2}{c}{UCF101 \cite{2012UCF101}} & \multicolumn{2}{c}{Adobe240fps \cite{su2017deep}} & \multicolumn{2}{c}{HMDB \cite{Kuehne11}} & \multicolumn{2}{c}{Vimeo \cite{xue2019video}} & \multicolumn{2}{c}{REDS \cite{Nah_2019_CVPR_Workshops_REDS}} & \multicolumn{2}{c}{Average}\\
                \cline{3-16}
                 &
                 & \makecell[c]{PSNR}
                 & \makecell[c]{SSIM}
                 & \makecell[c]{PSNR}
                 & \makecell[c]{SSIM}
                 & \makecell[c]{PSNR}
                 & \makecell[c]{SSIM}
                 & \makecell[c]{PSNR}
                 & \makecell[c]{SSIM}
                 & \makecell[c]{PSNR}
                 & \makecell[c]{SSIM}
                 & \makecell[c]{PSNR}
                 & \makecell[c]{SSIM}
                 & \makecell[c]{PSNR}
                 & \makecell[c]{SSIM}\\
                \hline

                \multirow{5}{*}{Specialized}& \makecell[c]{SepConv-Lf \cite{SimonNiklaus2}} & 31.68 & 0.9470 & 32.51 & 0.9473 & 36.36 & 0.9844 & 33.90 & 0.9483 & 33.49 & 0.9663 & 21.32 & 0.6965 & 31.54 & 0.9149\\
                \cline{3-16}
                &\makecell[c]{DAIN \cite{bao2019depth}} & 31.37 & 0.9452 & 32.72 & \textbf{0.9506} & 36.15 & 0.9837 & 33.89 & 0.9487 & 33.95 & \textbf{0.9701} & \textbf{21.85} & \textbf{0.7203} & \textbf{32.65}  & \textbf{0.9197}\\
                \cline{3-16}
                &\makecell[c]{CyclicGen \cite{liu2019cyclicgen}} & 31.54 & 0.9104 & 33.03 & 0.9303 & 35.60 & 0.9691 & 34.16 & 0.9242 & 33.04 & 0.9319 & 21.29 & 0.5686 & 31.44 & 0.8724\\
                \cline{3-16}
                &\makecell[c]{CyclicGen-large \cite{liu2019cyclicgen}} & 31.19 & 0.9004 & 32.43 & 0.9239 & 34.89 & 0.9626 & 33.75 & 0.9208 & 32.04 & 0.9163 & 20.92 & 0.5465 & 30.87 & 0.8617\\
                \cline{3-16}
                &\makecell[c]{BMBC \cite{park2020bmbc}} & 27.01 & 0.9223 & 27.92 & 0.9412 & 28.58 & 0.9569 & 28.42 & 0.9384 & 30.61 & 0.9629 & 21.21 & 0.7056 & 27.29 & 0.9045\\
                \cline{1-16}
                \multirow{2}{*}{General}&\makecell[c]{DVF \cite{liu2017video}} & 30.71 & 0.9303 & 32.31 & 0.9454 & 33.24 & 0.9613 & 33.59 & 0.9469 & 30.99 & 0.9379 & 20.44 & 0.6460 & 30.21 & 0.8946\\
                \cline{3-16}
                &\makecell[c]{Ours} & 31.44 & 0.9469 & 32.48 & 0.9504 & \textbf{36.42} & \textbf{0.9844} & 33.37 & \textbf{0.9492} & \textbf{34.03} & 0.9637 & 21.68 & 0.6959 & \textbf{31.57} & \textbf{0.9151}\\
                \cline{3-16}
    			\hline\hline
    		\end{tabular}
    	}
    \end{table*}

    \begin{figure*}[!h]
        \centering
    	\includegraphics[width=0.97\linewidth]{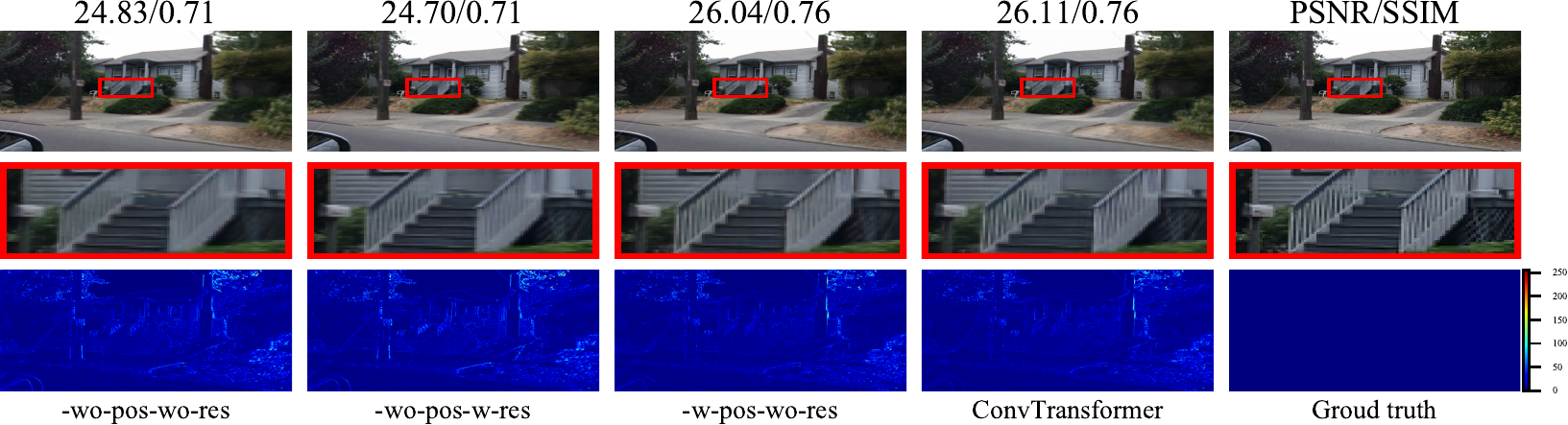}
        \caption{Effect of positional encoding and residual connection. The first three columns represent the visual results of ablation experiments. The second row represents the zoomed-in local details, while the third row shows the occlusion map with color bar displaying the pixel residual in the range of 0 to 255. The visual comparison in zoomed-in local details indicate that positional encoding module and reisudal connection is helpful for synthesizing photo-realistic images.}
    \end{figure*}

\subsection{Ablation Study}
    \begin{figure*}[]
        \centering
    	\includegraphics[width=0.93\linewidth]{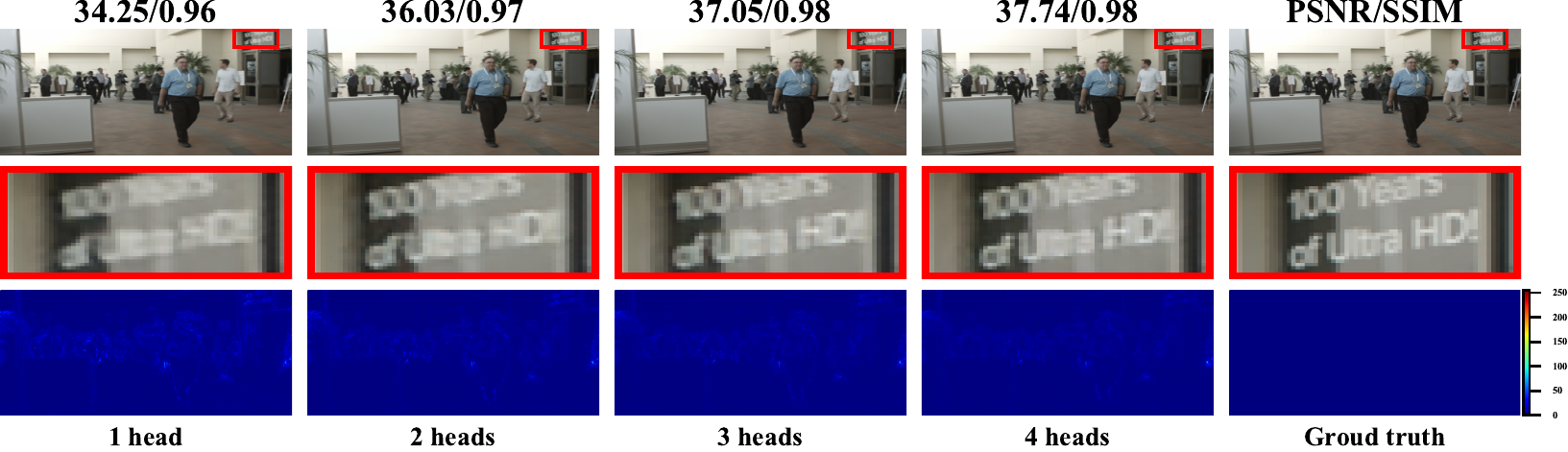}
    	\caption{Visual comparisons of ConvTransformer variants with different head numbers. The zoomed-in details are shown in the second row, and the third row illustrates the occlusion maps. The zoomed-in details demonstrate that ConvTransformer with more heads has an advantage in synthesizing details.}
    \end{figure*}

    In order to evaluate and justify the efficiency and superiority of each part in the proposed ConvTransformer architecture, several ablation experiments have been conducted in this work. Specifically, we gradually modify the baseline ConvTransformer model and compare their differences.
\subsubsection{Investigation of Positional Encoding and Residual Connection}
    \begin{table}
      \renewcommand{\arraystretch}{1.3}
      \centering
      \caption{Comparative results achieved with the ablation of residaul connection and positional encoding.}
      \resizebox{\linewidth}{!}{
        \begin{tabular}{c c c c c}
          \hline \hline
          Model & {Residual Connection} & {Positional Encoding} & PSNR & SSIM \\
          \hline
           ConvTransformer    &\Checkmark & \Checkmark & \textbf{29.8883} & \textbf{0.9383}\\
           ConvTransformer-wo-res &\XSolid    & \Checkmark & 29.6912 & 0.9367\\
           ConvTransformer-wo-pos &\Checkmark & \XSolid    & 28.4882 & 0.9173\\
           ConvTransformer-wo-res-wo-pos &\XSolid    &\XSolid & 28.4070 & 0.9104\\
          \hline \hline
        \end{tabular}
        }
    \end{table}
    We separately eliminate the positional encoding module, residual connection operation and both of them, and dub these three degradation networks as ConvTransformer-wo-pos, ConvTransformer-wo-res and ConvTransformer-wo-res-wo-pos, in which, the abbreviation wo represents without, pos indicates positional encoding and res is an abbreviation of residual connection. These three degradation algortihms are trained with the same trainset and implementation applied to ConvTransformer. We tabulate their performance on extrapolation task in terms of objective quantitative index PSNR and SSIM in Table 4, and visual comparisons in Figure 5.

    As summarized in Table 4, ConvTransformer has attained the best performance and achieves a relative performance gain of 0.1971dB, 1.4001dB and 1.4813dB in comparison with three degradation models ConvTransformer-wo-pos, ConvTransformer-wo-res and ConvTransformer-wo-res-wo-pos, respectively. Besides, visual comparisons in Figure 5 efficiently confirm these quantitative analysis. As shown in Figure 5, the stairs represented in zoomed-in area typically demonstrate that ConvTransformer with positional encoding and residual connection has an advance in suppressing artifacts and preserving local high-frequency details.
    \begin{figure}[]
        \centering
    	\includegraphics[width=0.93\linewidth]{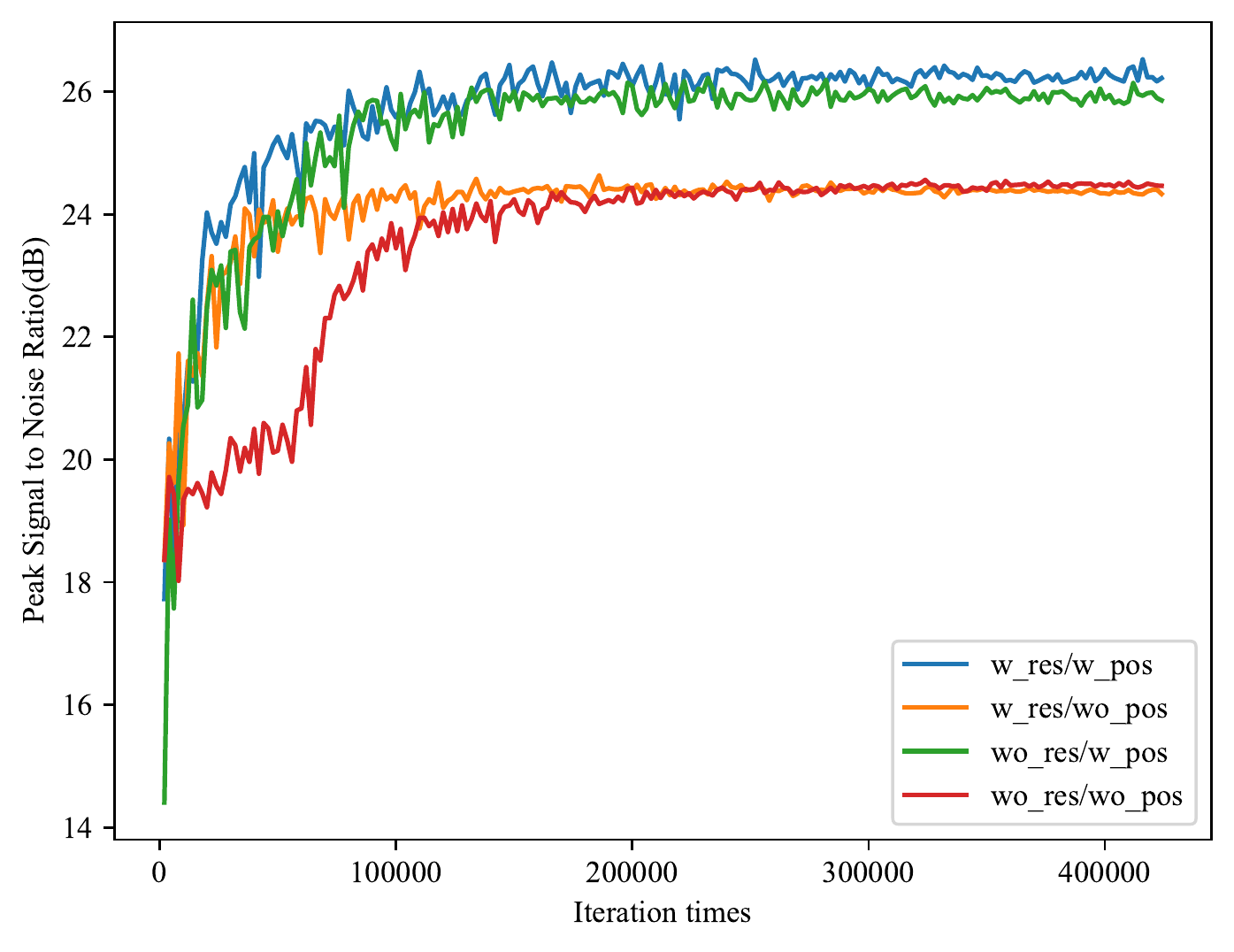}
    	\caption{Plots of PSNR convergence curves of ConvTransformer with the ablation of residual connection and positional encoding. w\_res/w\_pos represents ConvTransformer with residual connection and positional encoding, w\_res/wo\_pos represents ConvTranformer with the ablation of positional encoding, and wo\_res/w\_pos indicates that the residual connection is ablated.}
    \end{figure}

    \begin{figure}[]
        \centering
    	\includegraphics[width=0.93\linewidth]{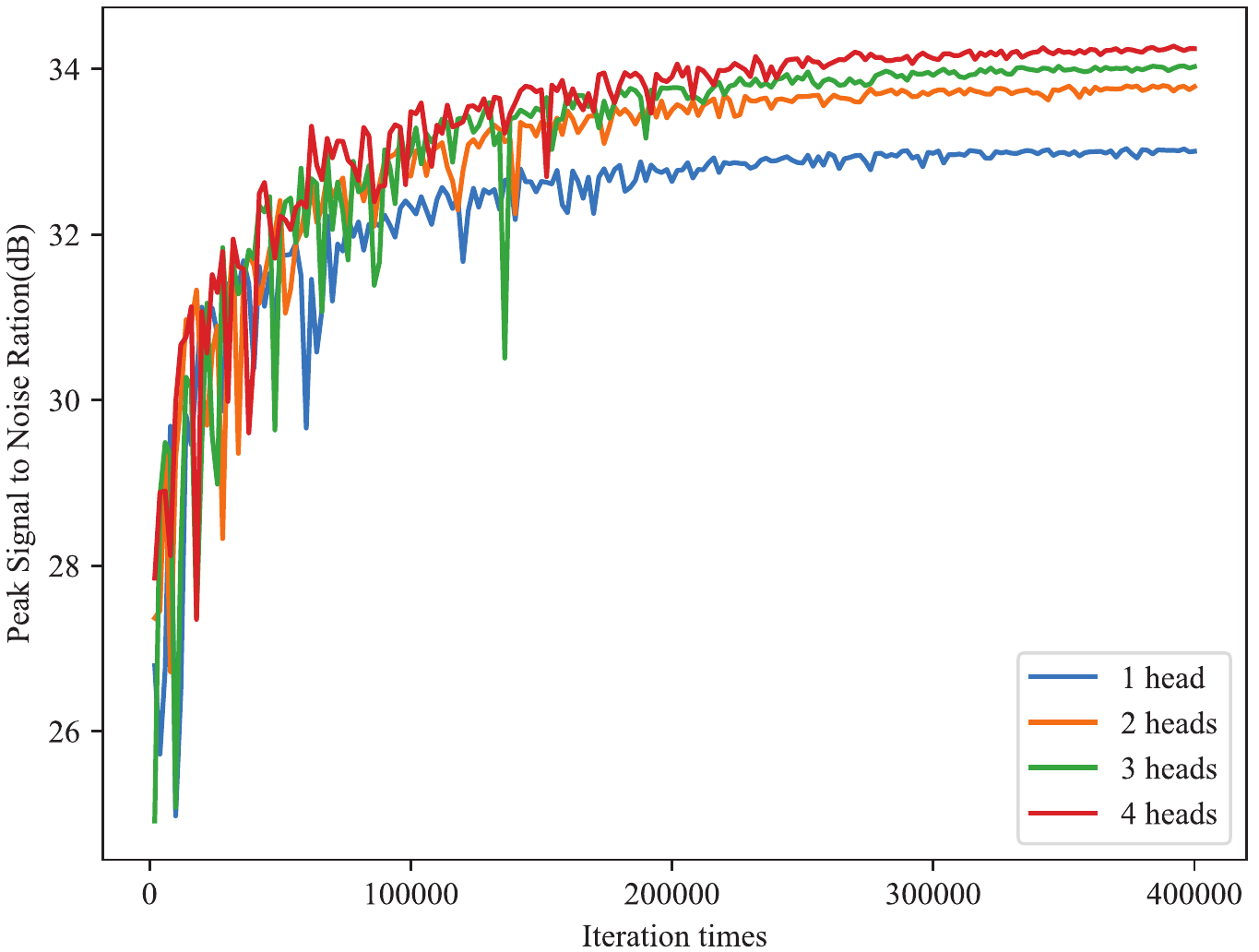}
    	\caption{Plots of PSNR convergence curves of ConvTransformer under different head numbers, i.e., 1 head, 2 heads, 3 heads and 4 heads.}
    \end{figure}

    We also visualize the convergence process of these three degradation networks in Figure 7. The convergence curves are consistent with the quantitative and qualitative comparisons above. As observed in Figure 7, we can find that the performance of ConvTransformer can easily be affected by the module positional encoding, and the residual connection can stabilize the training process.

    To sum up everything that has been stated so far, the positional encoding and residual connection architecture is beneficial for our proposed ConvTransformer to perform well.
\subsubsection{Investigation of Multi-Head Numbers Setting}

    \begin{figure*}[!h]
      \centering
        {\begin{overpic}[width=0.95\linewidth]{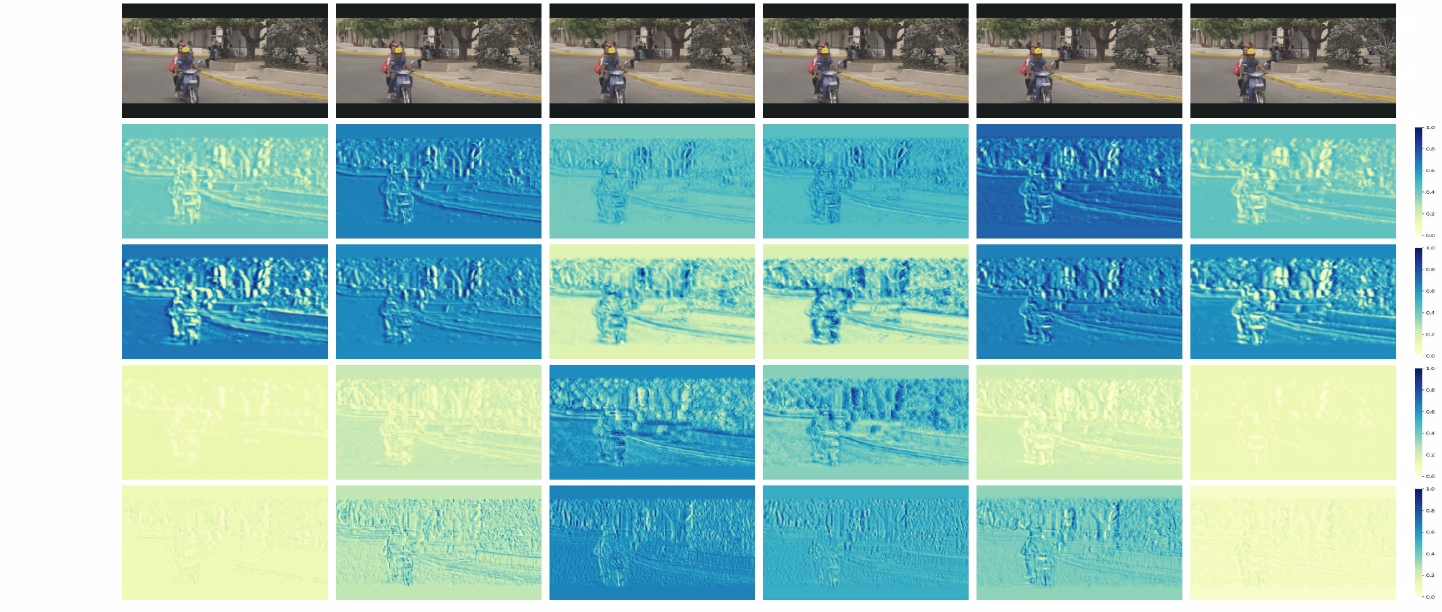}
            \put(15,43){\small\bfseries{$\mathcal{X}_0$}}
            \put(30,43){\small\bfseries{$\mathcal{X}_1$}}
            \put(44,43){\small\bfseries{$\mathcal{X}_2$}}
            \put(59,43){\small\bfseries{$\mathcal{X}_3$}}
            \put(74,43){\small\bfseries{$\mathcal{X}_4$}}
            \put(88,43){\small\bfseries{$\mathcal{X}_5$}}

            \put(1,29){\small\bfseries{${\rm Head}\_0$}}
            \put(1,20.5){\small\bfseries{${\rm Head}\_1$}}
            \put(1,12.5){\small\bfseries{${\rm Head}\_2$}}
            \put(1,4.5){\small\bfseries{${\rm Head}\_3$}}
        \end{overpic}} \hfill
      \caption{The attention maps $\mathcal{H}_{i}$ for decoding query $\mathcal{Q}_i$ on input sequence $\{\mathcal{X}_0, \mathcal{X}_1,...,\mathcal{X}_5\}$ under different heads. ColorBar with gradient from  yellow to blue represents the attention value in the range 0 to 1. The comparisons along each vertical column demonstrate that each head responds for exploiting different dependence. For example, the attention map $\mathcal{H}_{i-(1,3)}$, looking like a yellow map, indicates that frame $\mathcal{X}_3$ contributes local high-frequency information for target synthesize frame in Head\_1. Besides, the attention map $\mathcal{H}_{i-(2,2)}$, predominated by blue, represents that $\mathcal{X}_2$ supplies low-pass information for synthesizing target frame in Head\_2.}
    \end{figure*}

    The head numbers $H$ is a hyperparameter that allows ConvTransformer to jointly attend to information from different representation subspaces at different positions. In order to justify the efficiency of multi-head architecture, several multi-head variation experiments have been implemented in this work, and the quantitative index in terms of PSNR and SSIM, and visual examples are illustrated in Table 5 and Figure 6, respectively.

    As listed in Table 5, ConvTransformer-H-4, in the light of index PSNR, gains a relative performance gain of 1.2438dB in comparison with ConvTransformer-H-1. Besides, the visual comparisons, in Figure 6, indicate that ConvTransformer-H-4 generates more photo-realistic images. From these quantitative and visual comparisons, we can ascertain that more heads are helpful for ConTransformer architecture to incorporate information from different representation subspaces at different frame states.
    \begin{table}
        \renewcommand{\arraystretch}{1.5}
        \centering
        \caption{Comparative results achieved by ConvTransformer with different head numbers}
        \resizebox{\linewidth}{!}{
            \begin{tabular}{c c c c c}
              \hline \hline
              Model & ConvTransoformer-H-1 & ConvTransoformer-H-2 & ConvTransoformer-H-3 & ConvTransoformer-H-4 \\
              \hline
              PSNR & 33.0054 & 33.7861 & 34.0274  & \textbf{34.2438} \\
              SSIM & 0.9641  & 0.9695  & 0.9714   & \textbf{0.9731} \\
              \hline \hline
            \end{tabular}
        }
    \end{table}

    \begin{table}
        \renewcommand{\arraystretch}{1.5}
        \centering
        \caption{Quantitative evaluation of ConvTransformer with variation number of layers in Encoder and Decoder }\label{3}
        \resizebox{\linewidth}{!}{
            \begin{tabular}{c c c c c}
              \hline \hline
              Model & ConvTransoformer-L-1 & ConvTransoformer-L-2 & ConvTransoformer-L-3 & ConvTransoformer-L-5 \\
              \hline
              PSNR  & 34.2438 & 34.4623 & 34.4989  & \textbf{34.6031} \\
              SSIM  & 0.9731  & 0.9741  & 0.9744   & \textbf{0.9754} \\
              \hline \hline
            \end{tabular}
        }
    \end{table}

\subsubsection{Investigation of Layer Numbers Setting}
    The layer numbers $N$ is a hyperparameter which allows us to vary the capacity and computational cost of the encoder module and decoder module in ConvTransformer. To investigate the trade-off between performance and computational cost mediated by this hyperparameter, we conduct experiments with ConvTransformer for a range of different $N$ values. Such modified networks are named ConvTransformer-L-1, ConvTransformer-L-2, and so on. These layer variation networks were fully trained, and the performance indices of these networks are illustrated in Table 6.

    As tabulated in Table 6, we can easily find that the ConvTransformer-L-5 has acquired the optimal performance. In view of the PSNR index, ConvTransformer-L-5 has introduced a relative performance gain of 0.3593dB in comparison with ConvTransformer-L-1. Consequently, to achieve an excellent performance, it is desirable that the encoder and decoder contain more layers. However, more layers will take much more time for the networks to convergence, and consume more memory in training and testing. In summary, although more layers will improve the representative ability of ConvTransformer, we should set appropriate value $N$, in practice, and take the training time and memory burden into consideration.

\subsection{Long-Term Frame Sequence Dependence Analysis}
    In order to verify whether the proposed multi-head convolutional self-attention could efficiently capture the long-range sequential dependence within a video frame sequence, we visualize the attention maps of decoder query $\mathcal{Q}_i$ on input sequence $\mathcal{X}$ on decoder layer 1. As shown in Figure 9, the corresponding attention map $\mathcal{H}_{i-(k,j)}$ normalized in range [0, 1] represents the exploiting of input frame $\mathcal{X}_j$ for synthesizing the target frame $\mathcal{\hat Q}_i$ in head $k$. It should be emphasized that the attention value close to 1 is drawn with color blue, while the color yellow represents the attention value close to 0.

    With the vertical comparison in different positions, such as $\mathcal{X}_0$, $\mathcal{X}_1$, $\mathcal{X}_2$, $\mathcal{X}_3$, $\mathcal{X}_4$ and $\mathcal{X}_5$, we can find that different heads respond for incorporating different information for synthesizing target frames. For instance, attention map $\mathcal{H}_{i-(1,0)}$ predominant by blue indicates that attention layer in head 1 with position 0 mainly response for capturing the low-pass background dependence between frame $\mathcal{X}_0$ and target frame. Besides, along the lines of $\mathcal{H}_{i-(1,0)}$, the attention map $\mathcal{H}_{i-(1,2)}$ looking like a yellow map, revels that frame $\mathcal{X}_2$ supplies local high-frequency information for synthesizing target frame in head 1.


    In summary, through the analysis above and attention maps shown in Figure 9, a conclusion can be drawn that the proposed multi-head convolutional self-attention can efficiently model different long-term information dependencies, i.e., foreground information, background information and local high-frequency information.

\section{Conclusion}
    In this work, we propose a novel video frame synthesis architecture ConvTransformer, in which the multi-head convolutional self-attention is proposed to model the long-range spatially and temporally relation of frames in video sequence. Extensive quantitative and qualitative evaluations demonstrate that ConvTransfomer is a concise, compact and efficient model. The successful implementation of ConvTransformer sheds light on applying it to other video tasks that need to exploit the long-term sequential dependence in video frames.

{\small
\bibliographystyle{ieee_fullname}
\bibliography{egbib}
}

\clearpage
\appendix
\onecolumn
\centerline{\textbf{\huge Appendix}}
\renewcommand\thefigure{\arabic{figure}}
\renewcommand\thetable{\arabic{table}}
\section{Appendix-1: The details about module SFFN}

    \begin{figure*}[!ht]
    \centering
    \includegraphics[width=0.95\linewidth]{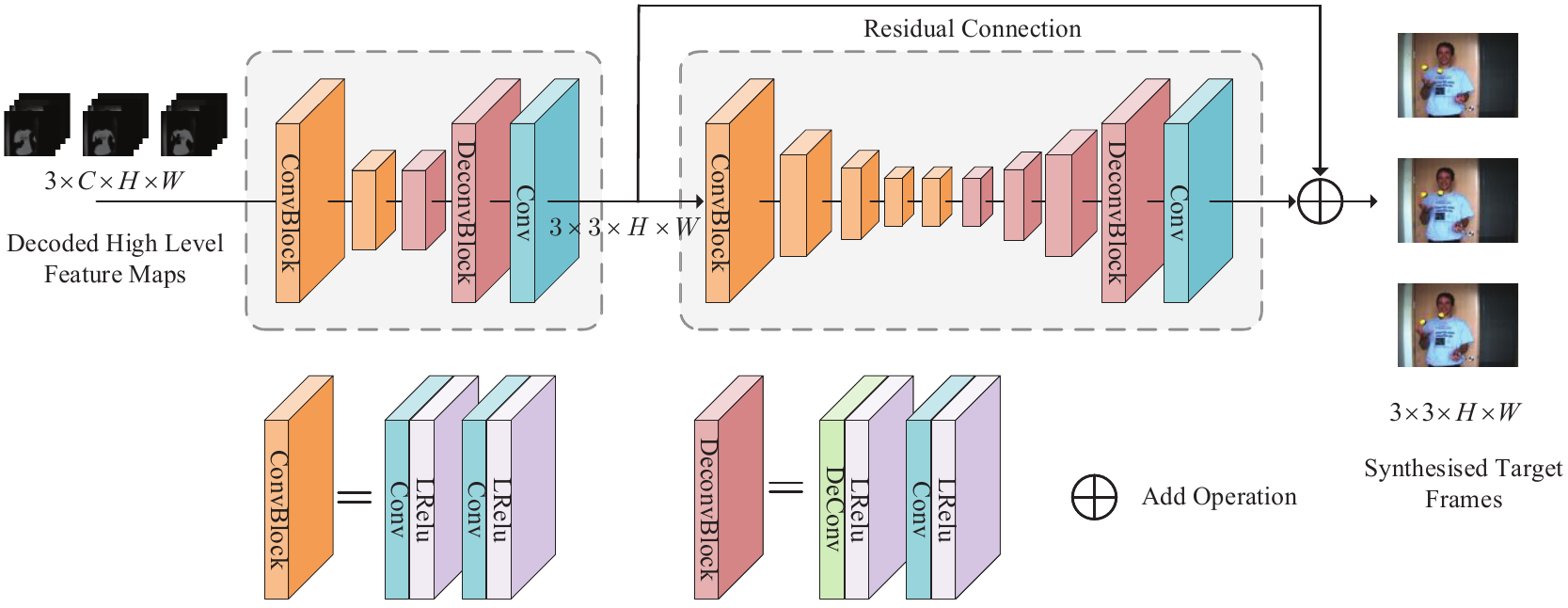}
    \caption{The design of module SFFN. SFFN consists of two U-Net like sub networks, i.e., SFFN1 and SFFN2. The pooling operation, for simplicity, and skip connection in U architecture are omitted. SFFN1 contains 5 layers, while SFFN2 is deeper, and includes 10 layers. The basic unit ConvBlock contains two convolutional layers, and DeconvBlock contains one deconvolutional layer and one convolutional layer. Given  decoded feature maps sequence $3*C*W*H$, the target synthesized frame sequence $3*3*H*W$ in RGB format is generated with the use of SFFN1 and SFFN2. A residual connection, in order to accelerate convergence, is built between the SFFN1 and the final output.}
    \end{figure*}

\section{Appendix-2: The prove of the effectiveness of Positional Map proposed in ConvTransformer}

    As shown in Figure 11, the position of each frame is first encoded with positional vector (PosVector) along the channel dimension, and then these PosVectors are expanded to the target positonal map (PosMap) through the repeating operation along the vertical (height) and horizontal (width) orientation. According to equations (3) and (4), for arbitrary pixel coordinate $(i,j)$ in frame $p+m$, the ${\rm PosMap}_{(p+m, (i,j))}$ can be represented as follows.

        \begin{figure}[!h]
            \begin{center}
                \includegraphics[width=0.7\linewidth]{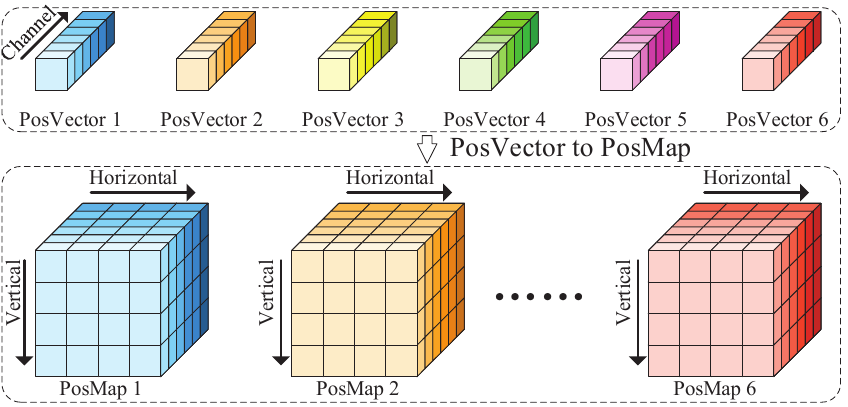}
            \end{center}
            \vspace{-1.2em}
            \caption{An illustration for positional map generating.}
            \vspace{-0.8em}
        \end{figure}

        \begin{equation}
          \resizebox{0.9\hsize}{!}{
          ${{\rm{Pos\_Ma}}{{\rm{p}}_{(p+m,(i,j,2k))}} = \sin ({w_k}(p+m))}= \sin({w_k}p)\cos({w_k}m) + \cos({w_k}p)\sin({w_k}m)$
          }
        \end{equation}

        \begin{equation}
          \resizebox{0.9\hsize}{!}{
          ${{\rm{Pos\_Ma}}{{\rm{p}}_{(p+m,(i,j,2k+1))}} = \cos ({w_k}(p+m))} = \cos({w_k}p)\cos({w_k}m) - \sin({w_k}p)\sin({w_k}m)$
          }
        \end{equation}

        \begin{equation}
          \resizebox{0.9\hsize}{!}{
          ${\left[
                \begin{array}{c}
                    {\rm PosMap}_{p+m,(i,j,2k)}\\
                    {\rm PosMap}_{p+m,(i,j,2k+1)} \\
                \end{array}
            \right] =
            \left[
                \begin{array}{cc}
                    \cos({w_k}m) & \sin({w_k}m)\\
                    -\sin({w_k}m) & \cos({w_k}m) \\
                \end{array}
            \right]
            \left[
                \begin{array}{c}
                     \sin({w_k}p)\\
                     \cos({w_k}p) \\
                \end{array}
            \right]}$
          }
        \end{equation}

        \begin{equation}
          \resizebox{.9\hsize}{!}{
          ${\left[
                \begin{array}{c}
                    {\rm PosMap}_{p+m,(i,j,2k)}\\
                    {\rm PosMap}_{p+m,(i,j,2k+1)} \\
                \end{array}
            \right] = \mathbf{M}
            \left[
                \begin{array}{c}
                     {\rm PosMap}_{p,(i,j,2k)} \\
                     {\rm PosMap}_{p,(i,j,2k+1)} \\
                \end{array}
            \right]}$
          }
        \end{equation}
    $w_k$ represents ${10000^{2k/{d_{model}}}}$. As illustrated in equations (14), (15), (16) and (17), the ${\rm PosMap}_{p+m}$ can be represented as a linear function of ${\rm PosMap}_{p}$, which proves that the proposed ${\rm PosMap}$ could efficiently model relative position relationship between different frames.

\section{Appendix-3:The parameters setting about ConvTransformer}
    Table 7 tabulates the parameter setting about one ConvTransformer. There are 4 heads, 7 encoder layers and 7 decoder layers.
    \begin{table}
      \centering
      \caption{The parameter setting of ConvTransformer for synthesizing 3 target frames. It includes 7 encoder layers and 7 decoder layers, while each attention layer of encoder and decoder consists of 4 heads. The $dmodel$ is 128, and hence, the depth of each head is 32. In each head, the $\rm Q\_Net$ generates the query feature maps, while the $\rm K\_V\_Net$ generates the key feature maps and value feature maps. }
      \resizebox{0.98\linewidth}{!}{
          \begin{tabular}{lll}
            \hline \hline
            \makecell[c]{Module} & \makecell[c]{Settings} & \makecell[c]{Output Size} \\
            \hline
            Input Sequence    & \makecell[c]{- - -} & $s \times 3 \times h \times w$ \\
            Feature Embedding & \makecell[c]{${\left[ \begin{array}{c} {{\rm Conv2D}\ {3 \times 3}\ s=1\ n=128}\end{array} \right]}\times 4$}
                              & $s \times 128 \times h \times w$ \\ \vspace{1ex}
            Encoder           & ${\left[ \begin{array}{c}
                                    {\begin{array}{cc}
                                           \begin{array}{c}
                                             {\rm Multi {\kern 2pt} Head {\kern 2pt} Convolutional {\kern 2pt} Self {\kern 2pt} Attention} \\
                                             (\rm Conducted {\kern 2pt} on {\kern 2pt} input {\kern 2pt} frame {\kern 2pt} sequence)
                                           \end{array} & {\left[ \begin{array}{c}
                                                   \vspace{0.2ex}
                                                   {\left[ \begin{array}{cc}
                                                              {\rm Head\_0} & {\left[ \begin{array}{cc}
                                                                                {\rm {Q\_Net\_Enc\_0}} & {\left[{{\rm Conv2D}\ {3 \times 3}\ s=1\ n=32} \right]} \\
                                                                                {\rm {K\_V\_Net\_Enc\_0}} & {\left[{{\rm Conv2D}\ {3 \times 3}\ s=1\ n=32} \right]} \\
                                                                                {\rm {Att\_Net\_Enc\_0}} & {\left[{{\rm Conv2D}\ {3 \times 3}\ s=1\ n=1} \right]}
                                                                              \end{array} \right]}
                                                            \end{array} \right]} \\
                                                   \vspace{0.2ex}
                                                   {\left[ \begin{array}{cc}
                                                              {\rm Head\_1} & {\left[ \begin{array}{cc}
                                                                                {\rm {Q\_Net\_Enc\_1}} & {\left[{{\rm Conv2D}\ {3 \times 3}\ s=1\ n=32} \right]} \\
                                                                                {\rm {K\_V\_Net\_Enc\_1}} & {\left[{{\rm Conv2D}\ {3 \times 3}\ s=1\ n=32} \right]} \\
                                                                                {\rm {Att\_Net\_Enc\_1}} & {\left[{{\rm Conv2D}\ {3 \times 3}\ s=1\ n=1} \right]}
                                                                              \end{array} \right]}
                                                            \end{array} \right]} \\

                                                   \vspace{0.2ex}
                                                   {\left[ \begin{array}{cc}
                                                              {\rm Head\_2} & {\left[ \begin{array}{cc}
                                                                                {\rm {Q\_Net\_Enc\_2}} & {\left[{{\rm Conv2D}\ {3 \times 3}\ s=1\ n=32} \right]} \\
                                                                                {\rm {K\_V\_Net\_Enc\_2}} & {\left[{{\rm Conv2D}\ {3 \times 3}\ s=1\ n=32} \right]} \\
                                                                                {\rm {Att\_Net\_Enc\_2}} & {\left[{{\rm Conv2D}\ {3 \times 3}\ s=1\ n=1} \right]}
                                                                              \end{array} \right]}
                                                            \end{array} \right]} \\
                                                   \vspace{0.2ex}
                                                   {\left[ \begin{array}{cc}
                                                              {\rm Head\_3} & {\left[ \begin{array}{cc}
                                                                                {\rm {Q\_Net\_Enc\_3}} & {\left[{{\rm Conv2D}\ {3 \times 3}\ s=1\ n=32} \right]} \\
                                                                                {\rm {K\_V\_Net\_Enc\_3}} & {\left[{{\rm Conv2D}\ {3 \times 3}\ s=1\ n=32} \right]} \\
                                                                                {\rm {Att\_Net\_Enc\_3}} & {\left[{{\rm Conv2D}\ {3 \times 3}\ s=1\ n=1} \right]}
                                                                              \end{array} \right]}
                                                            \end{array} \right]}
                                                 \end{array} \right]}
                                     \end{array}} \\
                                     \begin{array}{c}
                                        {\rm Feed {\kern 2pt} Forward} {\kern 150pt} {\left[ \begin{array}{c} {{\rm Conv2D}\ {3 \times 3}\ s=1\ n=128}\end{array} \right]}
                                     \end{array}
                                  \end{array} \right]}\times 7 {\kern 2pt}(number {\kern 2pt} of {\kern 2pt} encoding {\kern 2pt} layers)$    & $s \times 128 \times h \times w$ \\
            Query Sequence       & \makecell[c]{- - -} & $3 \times 128 \times h \times w$ \\
            Decoder           & ${\left[ \begin{array}{c}
                                    {\begin{array}{cc}
                                           {\kern 14pt} \begin{array}{c}
                                             {\rm Query {\kern 2pt} Self {\kern 2pt} Attention} \\
                                             (\rm  Conducted {\kern 2pt} on {\kern 2pt} query {\kern 2pt} frame {\kern 2pt} sequence)
                                           \end{array}   & {\left[ \begin{array}{c}
                                                   \vspace{0.2ex}
                                                   {\left[ \begin{array}{cc}
                                                              {\rm Head\_0} & {\left[ \begin{array}{cc}
                                                                                {\rm {Q\_Net\_Que\_0}} & {\left[{{\rm Conv2D}\ {3 \times 3}\ s=1\ n=32} \right]} \\
                                                                                {\rm {K\_V\_Net\_Que\_0}} & {\left[{{\rm Conv2D}\ {3 \times 3}\ s=1\ n=32} \right]} \\
                                                                                {\rm {Att\_Net\_Que\_0}} & {\left[{{\rm Conv2D}\ {3 \times 3}\ s=1\ n=1} \right]}
                                                                              \end{array} \right]}
                                                            \end{array} \right]} \\
                                                   \vspace{0.2ex}
                                                   {\left[ \begin{array}{cc}
                                                              {\rm Head\_1} & {\left[ \begin{array}{cc}
                                                                                {\rm {Q\_Net\_Que\_1}} & {\left[{{\rm Conv2D}\ {3 \times 3}\ s=1\ n=32} \right]} \\
                                                                                {\rm {K\_V\_Net\_Que\_1}} & {\left[{{\rm Conv2D}\ {3 \times 3}\ s=1\ n=32} \right]} \\
                                                                                {\rm {Att\_Net\_Que\_1}} & {\left[{{\rm Conv2D}\ {3 \times 3}\ s=1\ n=1} \right]}
                                                                              \end{array} \right]}
                                                            \end{array} \right]} \\

                                                   \vspace{0.2ex}
                                                   {\left[ \begin{array}{cc}
                                                              {\rm Head\_2} & {\left[ \begin{array}{cc}
                                                                                {\rm {Q\_Net\_Que\_2}} & {\left[{{\rm Conv2D}\ {3 \times 3}\ s=1\ n=32} \right]} \\
                                                                                {\rm {K\_V\_Net\_Que\_2}} & {\left[{{\rm Conv2D}\ {3 \times 3}\ s=1\ n=32} \right]} \\
                                                                                {\rm {Att\_Net\_Que\_2}} & {\left[{{\rm Conv2D}\ {3 \times 3}\ s=1\ n=1} \right]}
                                                                              \end{array} \right]}
                                                            \end{array} \right]} \\
                                                   \vspace{0.2ex}
                                                   {\left[ \begin{array}{cc}
                                                              {\rm Head\_3} & {\left[ \begin{array}{cc}
                                                                                {\rm {Q\_Net\_Que\_3}} & {\left[{{\rm Conv2D}\ {3 \times 3}\ s=1\ n=32} \right]} \\
                                                                                {\rm {K\_V\_Net\_Que\_3}} & {\left[{{\rm Conv2D}\ {3 \times 3}\ s=1\ n=32} \right]} \\
                                                                                {\rm {Att\_Net\_Que\_3}} & {\left[{{\rm Conv2D}\ {3 \times 3}\ s=1\ n=1} \right]}
                                                                              \end{array} \right]}
                                                            \end{array} \right]}
                                                 \end{array} \right]}
                                     \end{array}} \\
                                    {\begin{array}{cc}
                                           \begin{array}{c}
                                             {\rm Multi {\kern 2pt} Head {\kern 2pt} Convolutional {\kern 2pt} Self {\kern 2pt} Attention} \\
                                             {\rm (Conducted {\kern 2pt} on {\kern 2pt} query {\kern 2pt} frame {\kern 2pt} sequence}  \\
                                             {\rm and {\kern 2pt}encoded input {\kern 2pt} frame {\kern 2pt} sequence)}
                                           \end{array} & {\left[ \begin{array}{c}
                                                   \vspace{0.2ex}
                                                   {\left[ \begin{array}{cc}
                                                              {\rm Head\_0} & {\left[ \begin{array}{cc}
                                                                                {\rm {Q\_Net\_Dec\_0}} & {\left[{{\rm Conv2D}\ {3 \times 3}\ s=1\ n=32} \right]} \\
                                                                                {\rm {K\_V\_Net\_Dec\_0}} & {\left[{{\rm Conv2D}\ {3 \times 3}\ s=1\ n=32} \right]} \\
                                                                                {\rm {Att\_Net\_Dec\_0}} & {\left[{{\rm Conv2D}\ {3 \times 3}\ s=1\ n=1} \right]}
                                                                              \end{array} \right]}
                                                            \end{array} \right]} \\
                                                   \vspace{0.2ex}
                                                   {\left[ \begin{array}{cc}
                                                              {\rm Head\_1} & {\left[ \begin{array}{cc}
                                                                                {\rm {Q\_Net\_Dec\_1}} & {\left[{{\rm Conv2D}\ {3 \times 3}\ s=1\ n=32} \right]} \\
                                                                                {\rm {K\_V\_Net\_Dec\_1}} & {\left[{{\rm Conv2D}\ {3 \times 3}\ s=1\ n=32} \right]} \\
                                                                                {\rm {Att\_Net\_Dec\_1}} & {\left[{{\rm Conv2D}\ {3 \times 3}\ s=1\ n=1} \right]}
                                                                              \end{array} \right]}
                                                            \end{array} \right]} \\

                                                   \vspace{0.2ex}
                                                   {\left[ \begin{array}{cc}
                                                              {\rm Head\_2} & {\left[ \begin{array}{cc}
                                                                                {\rm {Q\_Net\_Dec\_2}} & {\left[{{\rm Conv2D}\ {3 \times 3}\ s=1\ n=32} \right]} \\
                                                                                {\rm {K\_V\_Net\_Dec\_2}} & {\left[{{\rm Conv2D}\ {3 \times 3}\ s=1\ n=32} \right]} \\
                                                                                {\rm {Att\_Net\_Dec\_2}} & {\left[{{\rm Conv2D}\ {3 \times 3}\ s=1\ n=1} \right]}
                                                                              \end{array} \right]}
                                                            \end{array} \right]} \\
                                                   \vspace{0.2ex}
                                                   {\left[ \begin{array}{cc}
                                                              {\rm Head\_3} & {\left[ \begin{array}{cc}
                                                                                {\rm {Q\_Net\_Dec\_3}} & {\left[{{\rm Conv2D}\ {3 \times 3}\ s=1\ n=32} \right]} \\
                                                                                {\rm {K\_V\_Net\_Dec\_3}} & {\left[{{\rm Conv2D}\ {3 \times 3}\ s=1\ n=32} \right]} \\
                                                                                {\rm {Att\_Net\_Dec\_3}} & {\left[{{\rm Conv2D}\ {3 \times 3}\ s=1\ n=1} \right]}
                                                                              \end{array} \right]}
                                                            \end{array} \right]}
                                                 \end{array} \right]}
                                     \end{array}} \\
                                     \begin{array}{c}
                                        {\rm Feed {\kern 2pt} Forward} {\kern 150pt} {\left[ \begin{array}{c} {{\rm Conv2D}\ {3 \times 3}\ s=1\ n=128}\end{array} \right]}
                                     \end{array}
                                  \end{array} \right]}\times 7 {\kern 2pt}(number {\kern 2pt} of {\kern 2pt} decoding {\kern 2pt} layers)$ & $s \times 128 \times h \times w$ \\
            Prediction        & $   \left[ {\begin{array}{cc}
                                    {\kern 90pt} {\rm SFFN} {\kern 75pt} & {\left[\begin{array}{ll}
                                       {\rm SFFN0} & \left[ \begin{array}{cc}
                                                              {\rm ConvBlock0} & \left[\begin{array}{c}
                                                                            {{{\rm Conv2D}\ {3 \times 3}\ s=1\ n=256} } \\
                                                                            {{{\rm Conv2D}\ {3 \times 3}\ s=1\ n=256} }
                                                                           \end{array} \right] \\
                                                              {\rm ConvBlock1} & \left[ \begin{array}{c}
                                                                            {{{\rm Conv2D}\ {3 \times 3}\ s=1\ n=512} } \\
                                                                            {{{\rm Conv2D}\ {3 \times 3}\ s=1\ n=512} }
                                                                           \end{array} \right] \\
                                                              {\rm DeconvBlock0} & \left[\begin{array}{c}
                                                                            {{{\rm Conv2D}\ {3 \times 3}\ s=1\ n=256} } \\
                                                                            {{{\rm Conv2D}\ {3 \times 3}\ s=1\ n=256} }
                                                                           \end{array} \right] \\
                                                              {\rm DeconvBlock1} & \left[\begin{array}{c}
                                                                            {{{\rm Conv2D}\ {3 \times 3}\ s=1\ n=128} } \\
                                                                            {{{\rm Conv2D}\ {3 \times 3}\ s=1\ n=128} }
                                                                           \end{array} \right]\\
                                                              {\rm Conv} & \left[\begin{array}{c}
                                                                            {{{\rm Conv2D}\ {3 \times 3}\ s=1\ n=64} } \\
                                                                            {{{\rm Conv2D}\ {3 \times 3}\ s=1\ n=32} } \\
                                                                            {{{\rm Conv2D}\ {1 \times 1}\ s=1\ n=3} }
                                                                           \end{array} \right]
                                                            \end{array} \right] \\
                                       {\rm SFFN1} & \left[ \begin{array}{cc}
                                                              {\rm ConvBlock0} & \left[\begin{array}{c}
                                                                            {{{\rm Conv2D}\ {3 \times 3}\ s=1\ n=32} } \\
                                                                            {{{\rm Conv2D}\ {3 \times 3}\ s=1\ n=32} }
                                                                           \end{array} \right] \\
                                                              {\rm ConvBlock1} & \left[\begin{array}{c}
                                                                            {{{\rm Conv2D}\ {3 \times 3}\ s=1\ n=64} } \\
                                                                            {{{\rm Conv2D}\ {3 \times 3}\ s=1\ n=64} }
                                                                           \end{array} \right] \\
                                                              {\rm ConvBlock2} & \left[\begin{array}{c}
                                                                            {{{\rm Conv2D}\ {3 \times 3}\ s=1\ n=128} } \\
                                                                            {{{\rm Conv2D}\ {3 \times 3}\ s=1\ n=128} }
                                                                           \end{array} \right] \\
                                                              {\rm ConvBlock3} & \left[\begin{array}{c}
                                                                            {{{\rm Conv2D}\ {3 \times 3}\ s=1\ n=256} } \\
                                                                            {{{\rm Conv2D}\ {3 \times 3}\ s=1\ n=256} }
                                                                           \end{array} \right] \\
                                                              {\rm ConvBlock4} & \left[\begin{array}{c}
                                                                            {{{\rm Conv2D}\ {3 \times 3}\ s=1\ n=512} } \\
                                                                            {{{\rm Conv2D}\ {3 \times 3}\ s=1\ n=512} }
                                                                           \end{array} \right] \\
                                                              {\rm DeconvBlock0} & \left[\begin{array}{c}
                                                                            {{{\rm Conv2D}\ {3 \times 3}\ s=1\ n=256} } \\
                                                                            {{{\rm Conv2D}\ {3 \times 3}\ s=1\ n=256} }
                                                                           \end{array} \right] \\
                                                              {\rm DeconvBlock1} & \left[\begin{array}{c}
                                                                            {{{\rm Conv2D}\ {3 \times 3}\ s=1\ n=128} } \\
                                                                            {{{\rm Conv2D}\ {3 \times 3}\ s=1\ n=128} }
                                                                           \end{array} \right] \\
                                                              {\rm DeconvBlock2} & \left[\begin{array}{c}
                                                                            {{{\rm Conv2D}\ {3 \times 3}\ s=1\ n=64} } \\
                                                                            {{{\rm Conv2D}\ {3 \times 3}\ s=1\ n=64} }
                                                                           \end{array} \right] \\
                                                              {\rm DeconvBlock3} & \left[\begin{array}{c}
                                                                            {{{\rm Conv2D}\ {3 \times 3}\ s=1\ n=32} } \\
                                                                            {{{\rm Conv2D}\ {3 \times 3}\ s=1\ n=32} }
                                                                           \end{array} \right] \\
                                                              {\rm Conv} & \left[\begin{array}{c}
                                                                            {{{\rm Conv2D}\ {1 \times 1}\ s=1\ n=3} }
                                                                           \end{array} \right]
                                                            \end{array} \right]
                                 \end{array} \right]}
                                 \end{array}}
                                 \right]$  & $s \times 3 \times h \times w$ \\
            Output Sequence & \makecell[c]{- - -} & $s \times 3 \times h \times w$ \\
            \hline
          \end{tabular}
      }
    \end{table}

\section{Appendix-4: Attention Calculation Process}
    The calculation process of multi-head convolutional self-attention in \textbf{Encoder}, query self-attention in \textbf{Decoder} and multi-head convolutional attention in \textbf{Decoder} are represented in Figure 12.
    \begin{figure*}[!ht]
    \centering
    \includegraphics[width=0.95\linewidth]{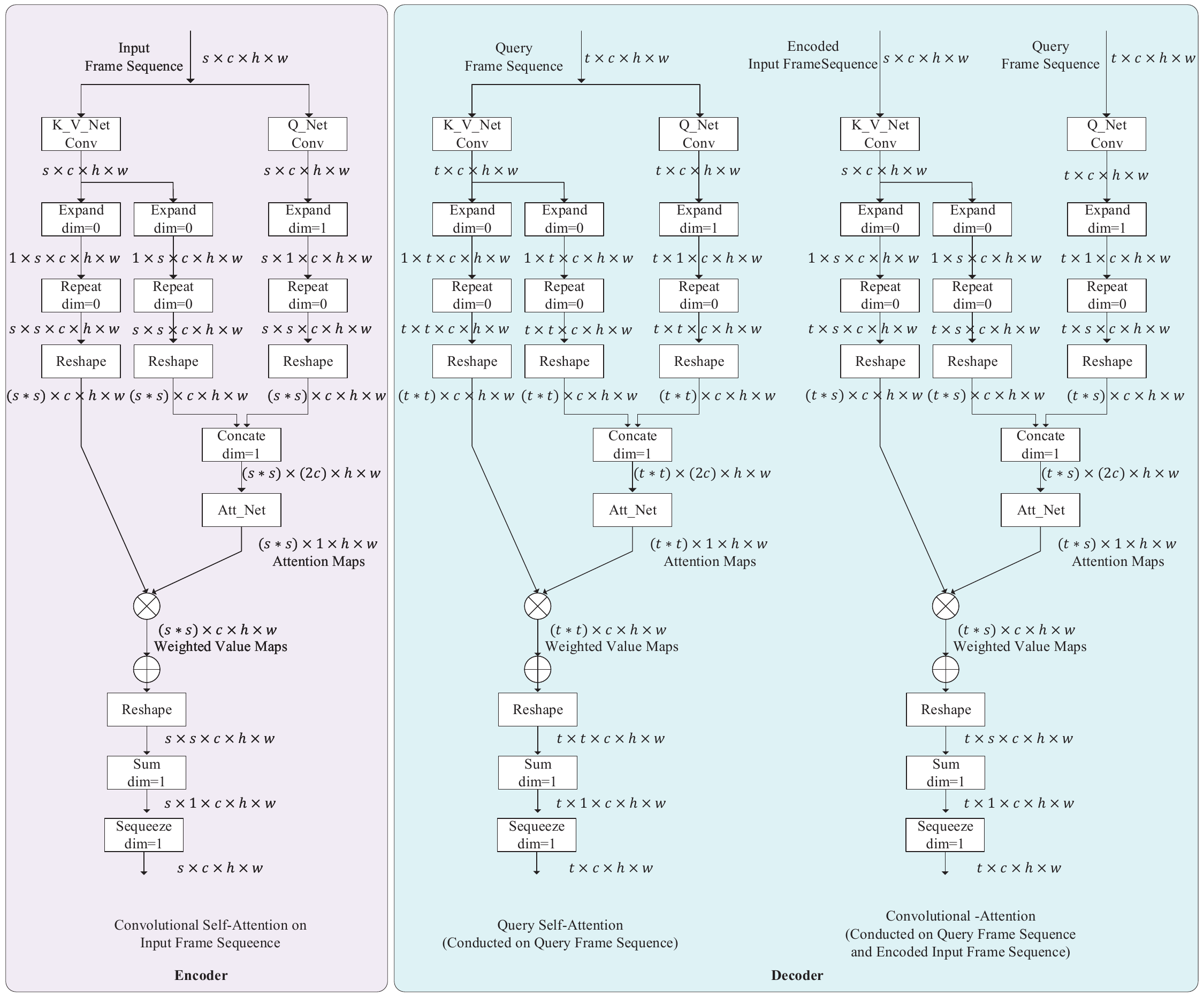}
    \caption{The calculation process of multi-head convolutional self-attention in Encoder, query self-attention in Decoder and multi-head convolutional attention in Decoder.}
    \end{figure*}

\end{document}